\documentclass{article}
\usepackage[utf8]{inputenc}
\usepackage{authblk}
\usepackage{setspace}
\usepackage[margin=1.25in]{geometry}
\usepackage{graphicx}
\graphicspath{ {./figures/} }
\usepackage{subcaption}
\usepackage{amsmath}
\usepackage{xcolor}
\usepackage{booktabs}

\title{Simultaneous Estimation of Manipulation Skill and Hand Grasp Force from Forearm Ultrasound Images}

\author[1*]{Keshav Bimbraw}
\author[2$\dag$]{Srikar Nekkanti}
\author[3]{Daniel B. Tiller II}
\author[1]{Mihir Deshmukh}
\author[1]{Berk Calli}
\author[4]{Robert D. Howe}
\author[1$\dag$]{Haichong K. Zhang}

\affil[1]{Robotics Engineering, Worcester Polytechnic Institute, Worcester, MA, USA.}
\affil[2]{Data Science, Worcester Polytechnic Institute, Worcester, MA, USA.}
\affil[$\dag$]{Biomedical Engineering, Worcester Polytechnic Institute, Worcester, MA, USA.}
\affil[3]{Inova Medical Group, Alexandria, VA, USA}
\affil[4]{Harvard Paulson School of Engineering and Applied Sciences, Cambridge, MA, USA}
\affil[*]{Address correspondence to: kbimbraw@wpi.edu}

\date{}

\begin{document}

\maketitle

%%%%%% Abstract %%%%%%
\begin{abstract}
% Final version from Kai 
% Berk
Accurate estimation of human hand configuration and the forces they exert is critical for effective teleoperation and skill transfer in robotic manipulation.
A deeper understanding of human interactions with objects can further enhance teleoperation performance. To address this need, researchers have explored methods to capture and translate human manipulation skills and applied forces to robotic systems. Among these, biosignal-based approaches, particularly those using forearm ultrasound data, have shown significant potential for estimating hand movements and finger forces. In this study, we present a method for simultaneously estimating manipulation skills and applied hand force using forearm ultrasound data. Data collected from seven participants were used to train deep learning models for classifying manipulation skills and estimating grasp force. Our models achieved an average classification accuracy of 94.87\%  $\pm$ 10.16\% for manipulation skills and an average root mean square error (RMSE) of 0.51 $\pm$ 0.19 N for force estimation, as evaluated using five-fold cross-validation.
These results highlight the effectiveness of forearm ultrasound in advancing human-machine interfacing and robotic teleoperation for complex manipulation tasks. This work enables new and effective possibilities for human-robot skill transfer and tele-manipulation, bridging the gap between human dexterity and robotic control.
%These results highlight the effectiveness of forearm ultrasound in advancing human-machine interfacing and robotic teleoperation for complex manipulation tasks. This work not only improves the accuracy of robotic manipulation but also paves the way for integrating human-like dexterity into robotic systems.

\end{abstract}

%%%%%% Main Text %%%%%%

\section{Introduction}
With the advances in robotics and artificial intelligence (AI), robots are increasingly being used to manipulate objects not only on the factory floor but also in everyday human environments, such as homes and healthcare settings \cite{dzedzickis2021advanced}.
Robotic manipulation finds applications in package handling \cite{surati2021pick}, manufacturing \cite{suomalainen2022survey}, healthcare \cite{bramhe2022robotic}, agriculture \cite{zhang2020state}, and space \cite{papadopoulos2021robotic} among others. 
In unstructured environments, robots must handle a diverse range of objects, requiring adaptable manipulation strategies to accomplish their tasks effectively.
Effectively interacting with different objects requires varied manipulation skills tailored to their unique properties and constraints \cite{thibaut2010developing}. 
This necessity arises due to the diverse geometries, textures, and mechanical properties of objects, which influence the selection of appropriate manipulation strategies \cite{cutkosky1990human}. For instance, flat objects may require sliding to an edge of a surface for getting access to grasping surfaces (for example, a credit card). 
If a cylindrical object is positioned against a vertical surface, a picking strategy could involve pushing it towards the surface to stabilize the grasp and facilitate access to suitable grasping areas, such as edges or protrusions. 
Alongside selecting the correct skill, applying the appropriate amount of force during interaction is equally crucial \cite{gao2005internal}. Insufficient force can lead to a loss of control, while excessive force might damage delicate objects or destabilize the task. Thus, a robust manipulation framework must integrate both skill identification and force estimation to achieve desired performance. These capabilities are essential for enabling robots to handle diverse objects and perform complex tasks with human-like adaptability. This is particularly relevant in applications such as autonomous manufacturing \cite{liu2022robot}, assistive technologies \cite{song2011force}, and robot-assisted surgery \cite{patel2022haptic} where task success depends on both the execution of appropriate skills and the transfer of the correct forces during object interaction. 

We propose an approach for simultaneously estimating manipulation skills and hand forces using ultrasound data from the forearm. By leveraging brightness mode (B-mode) ultrasound, our method addresses a central requirement shared by two primary paradigms in human-like robotic manipulation: teleoperation and Learning from Demonstrations (LfD). Both of these approaches rely on accurate skill identification and precise force estimation.

Teleoperation mirrors the human operator's movements in real time and is often employed in scenarios where precise, immediate control is necessary, such as remote surgical systems \cite{haidegger2011surgery} or hazardous environment operations \cite{trevelyan2016robotics}. Although teleoperation is limited by communication latency and scalability issues, it remains highly useful for tasks that require continuous human oversight \cite{lichiardopol2007survey}. Wearable gloves and vision-based techniques have been used to teleoperate robotic systems \cite{dipietro2008survey, pisharady2015recent}. However, gloves restrict natural hand movements, while vision-based systems suffer from occlusion, poor lighting conditions, and shadow artifacts.

Bodily signals have been used as an alternative, with surface electromyography (sEMG) \cite{zheng2022surface, bimbraw2023towards}, force myography (FMG) \cite{cho2016force, bimbraw2024random}, near-infrared spectroscopy (NIRS) \cite{tsubone2007application}, electric impedance tomography (EIT) \cite{zheng2020electrical}, and ultrasound \cite{yang2024ultrasound, he2024ultrasound} all explored for robotic control. Surface electromyography suffers from inherent limitations such as high sensitivity to electrode location and issues with electrode–skin contact \cite{zheng2022surface}. FMG faces susceptibility to sensor drift and signal variability due to preload inconsistencies \cite{xiao2019review}, while NIRS is sensitive to tissue thickness and ambient light interference \cite{tchantchane2023review}. EIT encounters challenges in image reconstruction that impact measurement sensitivity, range, and response time \cite{cui2023recent}. Ultrasound, although limited by sensor size, has shown promise for estimating hand gestures \cite{bimbraw2020towards}, finger angles \cite{bimbraw2023simultaneous}, and finger forces \cite{bimbraw2023estimating}. Furthermore, ongoing efforts are focused on miniaturizing ultrasound sensors \cite{spacone2024tracking} and making them more wearable \cite{bimbraw2024mirror} for human–machine interfacing applications.

LfD enables robots to learn task constraints and execution strategies by observing expert demonstrations \cite{schaal1996learning}. Using supervised learning, LfD aligns closely with tasks where ideal behavior is difficult to formalize but can be effectively demonstrated \cite{ravichandar2020recent}. This approach has been extensively explored for trajectory learning, policy optimization, and skill generalization in applications ranging from manufacturing to healthcare \cite{liang2020teaching, su2021toward}. Multiple sensor data acquisition is commonly utilized to encode task-specific variables effectively. For instance, \cite{zeng2019encoding} integrated multimodal data (motion trajectories, stiffness profiles obtained via sEMG, and force/torque measurements) to capture correlations between position, velocity, stiffness, and force, allowing precise reproduction of tasks with combined position and force constraints. Similarly, Vasan et al. demonstrated the application of LfD to train sEMG-based prostheses using contralateral limb movements \cite{vasan2017learning}. By integrating sEMG signals and joint trajectories from the intact limb, they mapped user demonstrations into a policy for controlling multi degree-of-freedom prosthetic devices.

The choice between teleoperation and LfD depends on task-specific requirements. Teleoperation is advantageous for high-stakes, dynamic tasks requiring direct human intervention, whereas LfD offers scalability and autonomy by enabling robots to adapt and generalize behaviors based on prior demonstrations. Both teleoperation and LfD demand robust skill identification and force sensing, motivating our approach. Our proposed ultrasound-based approach supports both paradigms by providing real-time skill and force estimation (for teleoperation) and serving as a rich source of task demonstrations (for LfD). By focusing on motion primitives underlying manipulation tasks, rather than strict trajectory replication, we facilitate efficient transfer of human-like dexterity to robots. This capability enhances the adaptability and precision of robotic manipulation systems in dynamic and unstructured environments.

\subsection{Related Works}
Zongxing et al. provided a comprehensive review of modalities for simultaneous gesture recognition and force assessment, including vision-based systems, surface electromyography (sEMG), ultrasound imaging, and hybrid approaches \cite{zongxing2023human}. Most of the modalities focus on discrete force levels and gesture recognition for human-machine interaction systems. In contrast, our work focuses on combining discrete manipulation skill classification with continuous force estimation, addressing challenges specific to robotic control in industrial environments. Using sEMG, Hu et al. proposed a myoelectric control framework for synchronous gesture recognition and force estimation \cite{hu2022novel}. Their work improves robustness under variable force conditions, targeting applications in prosthetics and exoskeletons, estimating discrete force-levels for wearable assistive technologies. Our approach diverges by using ultrasound imaging to classify discrete manipulation skills and estimate continuous force values aimed at robot control. 

Zengyu et al. introduced a two-stage cascade model utilizing A-mode ultrasound for gesture classification and force estimation \cite{zengyu2022simultaneous}. Their method focuses on discrete force levels and gestures for human-machine interfaces. In contrast, we employ B-mode ultrasound imaging to integrate manipulation skill recognition with continuous force prediction, specifically designed for precision in industrial robotic manipulation tasks. Peng et al. developed a flexible, wearable B-mode ultrasound transducer for simultaneous recognition of hand movements and force levels, achieving high accuracy in both amputees and non-disabled individuals \cite{peng2023novel}. Although their work focuses on discrete classifications of movements and force levels for prosthetic applications, our approach advances these capabilities by utilizing B-mode ultrasound imaging for discrete manipulation skill recognition and continuous force estimation. This enables adaptive robotic control while simultaneously capturing force and skill data to facilitate more efficient LfD.

\subsection{Contributions}
Given the limitations of current robotic teleoperation systems and models trained with LfD approaches in accurately replicating complex human manipulation skills and force control, as highlighted in \cite{kroemer2021review, della2019learning}, our work addresses these gaps through the following contributions.
\begin{enumerate}
    \item \textbf{Manipulation Skill Classification}: We propose a deep learning model capable of classifying multiple manipulation skills derived from a robotic manipulation dataset using ultrasound imaging of the forearm. These skills encompass a wide range of real-world object manipulation. Testing with seven human subjects performing five distinct manipulation tasks yielded a five-fold cross-validated classification accuracy of 94.9 ± 10.2\%.

    \item \textbf{Continuous Force Estimation}: We predict applied forces during manipulation tasks with high accuracy across multiple subjects. An average force estimation RMSE of 0.51 ± 0.19 N was obtained.

    \item \textbf{Joint Skill and Force Estimation}: We demonstrate that ultrasound data from the forearm can effectively estimate both manipulation skills (post-object contact) and the continuous force exerted by the human hand on the manipulated object. We provide a video to demonstrate the performance.

    \item \textbf{Interpretability and Robustness}: The proposed CNN-based framework uses Grad-CAM visualizations to interpret the regions of ultrasound images most relevant for predicting manipulation skills and forces. These heatmaps highlighted key muscle groups, such as the flexor digitorum profundus (FDP) and flexor pollicis longus (FPL), and demonstrated the model’s ability to handle inter-subject variability and align predictions with known muscle functions.
\end{enumerate}

To summarize, this paper explores a novel framework for simultaneously estimating manipulation skills and forces from forearm ultrasound data. Testing with seven human subjects performing five distinct manipulation tasks yielded a five-fold cross-validated classification accuracy of 94.9 ± 10.2\% and an average force estimation RMSE of 0.51 ± 0.19 N. These findings highlight the feasibility of ultrasound-based sensing for robust teleoperation and LfD, where accurate skill identification and precise force profiling enable human-like dexterity in real-world scenarios. Section~\ref{sec:methods} describes the data acquisition setup, manipulation skill classification, and force estimation methodologies. The experimental results and analysis are presented in Section~\ref{sec:results}, including Grad-CAM visualizations for model interpretability. Finally, Section~\ref{sec:discussion} discusses the implications of the findings, limitations, and future research directions.

\section{Materials and Methods}\label{sec:methods}
Robust robotic manipulation requires precise motion planning and the ability to execute skills with task-specific force dynamics. Inspired by prior work on skill learning and motion primitives~\cite{della2019learning, eppner2015exploitation}, this study proposes a framework integrating manipulation skill classification and force estimation using ultrasound imaging of forearm muscles. 

\begin{figure}[ht]
    \includegraphics[width=\textwidth]{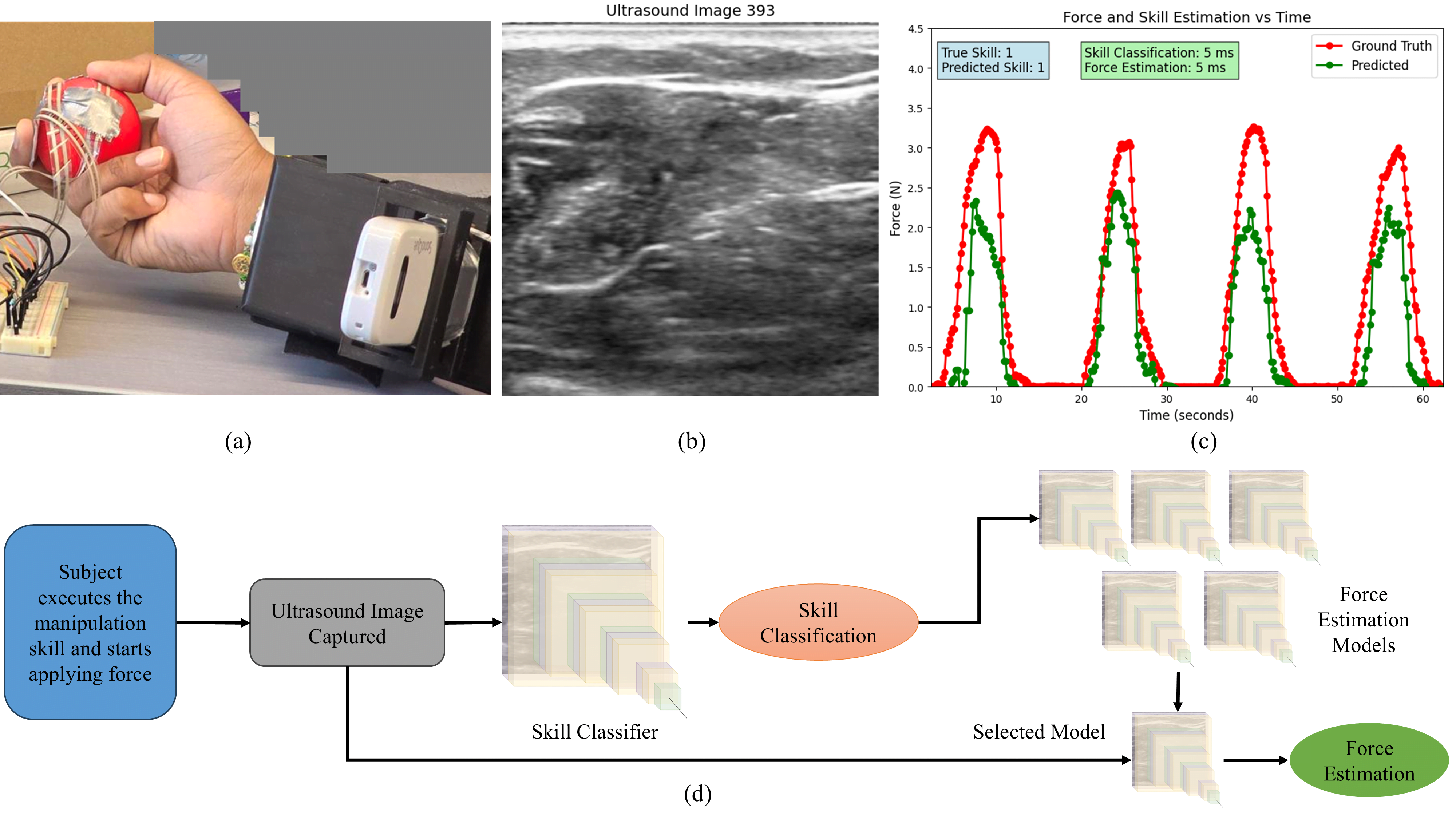}
    \caption{The pipeline for simultaneous manipulation skill classification and continuous force estimation using ultrasound data and deep learning models. (a) A subject performing manipulation skill 1 and applying force to a ball. (b) Forearm ultrasound image highlighting muscle activity. (c) Real-time force and skill estimation compared to ground truth. (d) Pipeline architecture showing the integrated skill classification and force estimation for robotic control.}
\label{fig:pipeline}
\end{figure}

\subsection{Pipeline}
The goal of this pipeline is to enable real-time manipulation skill classification and continuous force estimation using ultrasound imaging of the forearm. This has applications in in robotic teleoperation and Learning from Demonstration (LfD) frameworks. Figure \ref{fig:pipeline} shows the different components of the pipeline. Our approach involves four stages: ultrasound data is first acquired from subjects performing predefined manipulation tasks designed to replicate real-world scenarios. Next, the data is processed through a custom deep learning framework to classify manipulation skills and estimate corresponding forces. The estimated skills and forces can then be passed on for downstream robotic control, or for training LfD frameworks. The pipeline, shown in Figure \ref{fig:pipeline}(d), summarizes this process.

As the subjects execute predefined manipulation tasks as shown in Figure \ref{fig:pipeline}(a), ultrasound data from the forearm is acquired to capture forearm muscle activity. The ultrasound data can be visualized in Figure \ref{fig:pipeline}(b). The acquired ultrasound image frame is first processed using a deep learning model to classify manipulation skills. This step identifies discrete manipulation types based on the muscle activation patterns observed in the ultrasound images. Once the manipulation skill is identified, a corresponding pre-trained model is used to estimate the force applied by the hand as shown in Figure \ref{fig:pipeline}(d).

The classified manipulation skills and estimated forces are simultaneously obtained to support robotic manipulation. This can be seen in Figure \ref{fig:pipeline}(c). For teleoperation, this integration can provide real-time guidance for robot control, enabling intuitive and precise task execution. For LfD, the estimated skill and force data can be used to train models capable of autonomous manipulation, allowing robots to adapt to unstructured and dynamic environments efficiently.

\subsection{Manipulation Skills}
\begin{figure}[ht]
    \centering
    % Top row: Original skill prototypes
    \begin{subfigure}{0.19\textwidth}
        \includegraphics[width=\textwidth]{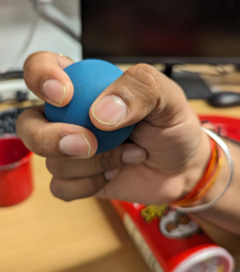}
        \caption{Subject Skill 1}
    \end{subfigure}
    \begin{subfigure}{0.19\textwidth}
        \includegraphics[width=\textwidth]{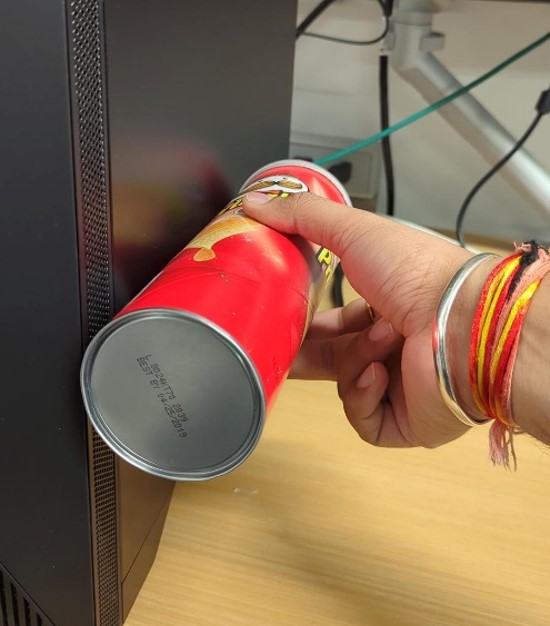}
        \caption{Subject Skill 2}
    \end{subfigure}
    \begin{subfigure}{0.19\textwidth}
        \includegraphics[width=\textwidth]{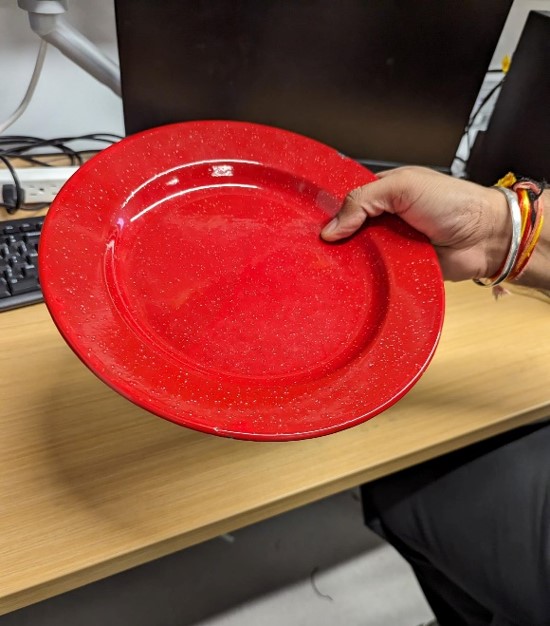}
        \caption{Subject Skill 3}
    \end{subfigure}
    \begin{subfigure}{0.19\textwidth}
        \includegraphics[width=\textwidth]{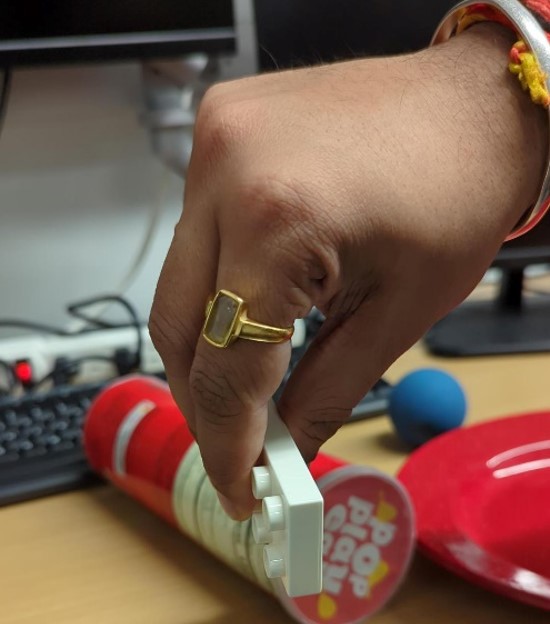}
        \caption{Subject Skill 4}
    \end{subfigure}
    \begin{subfigure}{0.19\textwidth}
        \includegraphics[width=\textwidth]{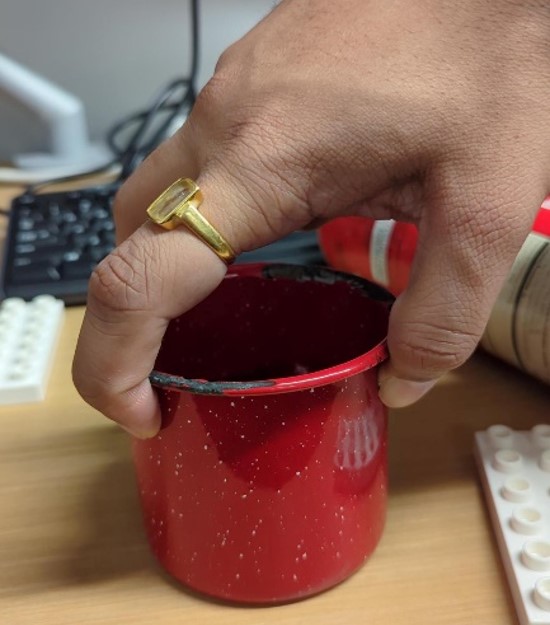}
        \caption{Subject Skill 5}
    \end{subfigure}

    \vspace{1em} % Add some vertical space between the two rows

    % Bottom row: Robot executing the skills
    \begin{subfigure}{0.19\textwidth}
        \includegraphics[width=\textwidth]{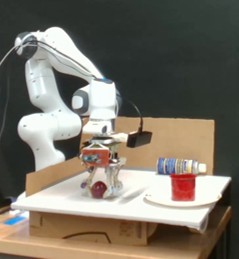}
        \caption{Robot Skill 1}
    \end{subfigure}
    \begin{subfigure}{0.19\textwidth}
        \includegraphics[width=\textwidth]{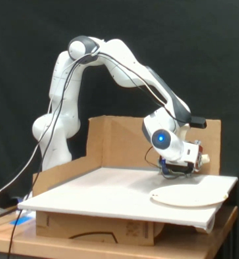}
        \caption{Robot Skill 2}
    \end{subfigure}
    \begin{subfigure}{0.19\textwidth}
        \includegraphics[width=\textwidth]{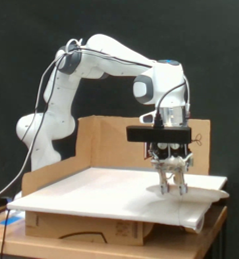}
        \caption{Robot Skill 3}
    \end{subfigure}
    \begin{subfigure}{0.19\textwidth}
        \includegraphics[width=\textwidth]{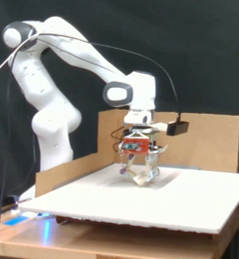}
        \caption{Robot Skill 4}
    \end{subfigure}
    \begin{subfigure}{0.19\textwidth}
        \includegraphics[width=\textwidth]{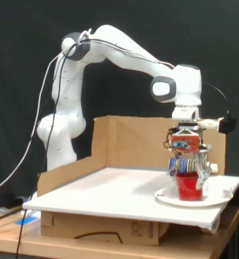}
        \caption{Robot Skill 5}
    \end{subfigure}

    \caption{(Top row) Manipulation skill prototypes and objects: (a) Push to horizontal - ball, (b) Push to vertical - cylindrical can, (c) Slide to edge - plate, (d) Flip - thin cuboid, (e) Simple pick (Push grasp) - mug. (Bottom row) Robot executing the corresponding manipulation skills (f - j).}
    \label{fig:composite}
\end{figure}

The selected manipulation skills are Push-to-Horizontal, Push-to-Vertical, Slide-to-Edge, Flip, and Simple-Pick. They address common challenges in handling diverse objects with varied shapes and stability requirements. For instance, Push-to-Horizontal, following \cite{eppner2015exploitation} is advantageous for objects prone to rolling, such as balls, as it guides the gripper’s fingers to stabilize the object through sweeping all three fingers from all sides trapping the object and stabilizing the grasp. Similarly, Push-to-Vertical, as also noted by \cite{eppner2015exploitation}, leverages pushing against vertical barriers, stabilizing items such as bottles. The Slide-to-Edge skill, discussed in \cite{eppner2015exploitation}, enables robots to handle flat objects like plates by sliding them to an edge, replicating a human technique that improves grasp stability.  Following \cite{odhner2012precision} and \cite{odhner2013open}’s human dexterity studies, we have Flip for facilitating small, flat objects like coins, where one finger supports the object while another lifts it.  For objects that don’t require dexterous strategies, Simple-Pick allows straightforward handling without additional manipulation. The different skills are shown in Figure \ref{fig:composite}.

In this work, we classify ultrasound data from the forearm to estimate manipulation skill. For each skill, these is a distinctive signature corresponding to each motion primitive observed upon contact in direct human demonstrations. While studies like \cite{manschitz2020learning} show how human-guided robot actions can classify force-intensive tasks, and \cite{gao2019learning} highlights human-inspired learning by recording robot movement which incorporates human guidance, our approach enables force prediction and skill classification directly from human arm muscle movement. Our method, using ultrasound imaging, allows the robot to apply the most suitable skill with a response that mirrors human-like dexterity, enabling more intuitive teleoperation as well as obtaining data for LfD.

\subsection{Data Acquisition}
To ensure consistent and high-quality data collection, both ultrasound and force data were acquired during manipulation tasks. Per skill and subject, 2000 frames of ultrasound data were obtained. The following subsections describe the system setup and ground truth collection methods.
\subsubsection{System Setup}
The ultrasound probe (Sonostar 5L linear palm Doppler medical ultrasound probe) was secured to the subject’s forearm using Velcro straps within a custom-designed 3D-printed casing. The probe was positioned and adjusted to maximize image quality. This probe employs an electronic linear array scanning mode, operating at 7.5/10 MHz with 128 elements and an adjustable depth range of 20–55 mm. Ultrasound images, captured at 500 × 500 pixels resolution and an average frame rate of 6.3 Hz, were displayed and configured via a Windows interface. For 2000 frames, this meant about 5 minutes and 18 seconds of data acquisition time on an average.

Force data were collected using three FlexiForce A201 sensors (9.7 mm sensing area, 4 N range) attached to five distinct objects, each corresponding to a different manipulation skill. The sensors measured the force applied by the fingers during the tasks, capturing fine-grained force variations.

The capture of ultrasound frames was aligned with the force data collected from the sensors, ensuring accurate temporal alignment between the two datasets. To manage this synchronization, an Arduino Uno microcontroller was used to interface with the force sensors, and a custom Python script was used to acquire both ultrasound screenshots and the data from the Arduino interfaced force sensors. 

\subsubsection{Ground Truth}
For acquiring the data for a particular skill, the subjects were instructed to hold the objects with thumb, index and middle fingers. Since each object involved a different manipulation skill, per subject, each session was labelled with the skill being executed. 
For continuous force estimation, the subjects were instructed to apply varying levels of force on the objects. To ensure consistent application and modulation of force during manipulation tasks, subjects were guided by a structured sequence of auditory cues that corresponded to specific frames within each 100-frame segment of the dataset. These cues helped subjects synchronize their actions with data acquisition:

\begin{itemize}
    \item \textbf{Start Applying Force:} At the 20th frame, a 500 Hz beep lasting 250 milliseconds was played, prompting the subject to initiate force application gradually.
    \item \textbf{Hold Maximum Force:} At the 40th frame, a 750 Hz beep lasting 250 milliseconds instructed the subject to reach and hold maximum force.
    \item \textbf{Start Reducing Force:} At the 60th frame, a 600 Hz beep lasting 250 milliseconds indicated that the subject should begin reducing the applied force.
    \item \textbf{Stop Applying Force:} At the 80th frame, a 550 Hz beep lasting 250 milliseconds signaled the subject to stop applying force entirely.
\end{itemize}

This structured timing of auditory signals was repeated for every 100 frames of data acquisition, ensuring consistent ground truth force profiles across subjects and skills. 

\subsection{Human Subjects and IRB Approval}
The study was approved by the institutional research ethics committee under protocol number IRB-23-0634, and written informed consent was obtained from all subjects prior to the commencement of the sessions. Ultrasound data was captured from the forearm of choice for 7 subjects (5 males, 2 females; Age: 27.9 $\pm$ 3.4 years; Height: 172.1 $\pm$ 8.8 cm; diameter of the forearm around the point where the probe was placed: 18.9 $\pm$ 2.9 cm). Table \ref{table_1} lists the sex and forearm diameter at the ultrasound probe location for each subject enrolled in the study.

\begin{table}[h!]
    \centering
    \begin{tabular}{|l|c|c|c|c|c|c|c|}
        \hline
        & \textbf{TS1} & \textbf{TS2} & \textbf{TS3} & \textbf{TS4} & \textbf{TS5} & \textbf{TS6} & \textbf{TS7} \\ \hline
        \textbf{Age (yr)} & 29 & 23 & 30 & 33 & 30 & 24 & 26 \\ \hline
        \textbf{Sex (M/F)} & M & M & M & M & M & F & F \\ \hline
        \textbf{Height (cm)} & 180 & 169 & 175 & 175 & 185 & 158 & 163 \\ \hline
        \textbf{Forearm Dia. (cm)} & 25 & 20 & 17 & 20 & 17 & 17 & 16 \\ \hline
    \end{tabular}
    \caption{Demographic and Forearm Measurements of Test Subjects}
    \label{table_1}
\end{table}

\subsection{Model Architecture}
\begin{figure}[ht!]
    \centering
    \begin{subfigure}[t]{0.486\textwidth}
        \centering
        \includegraphics[width=\textwidth]{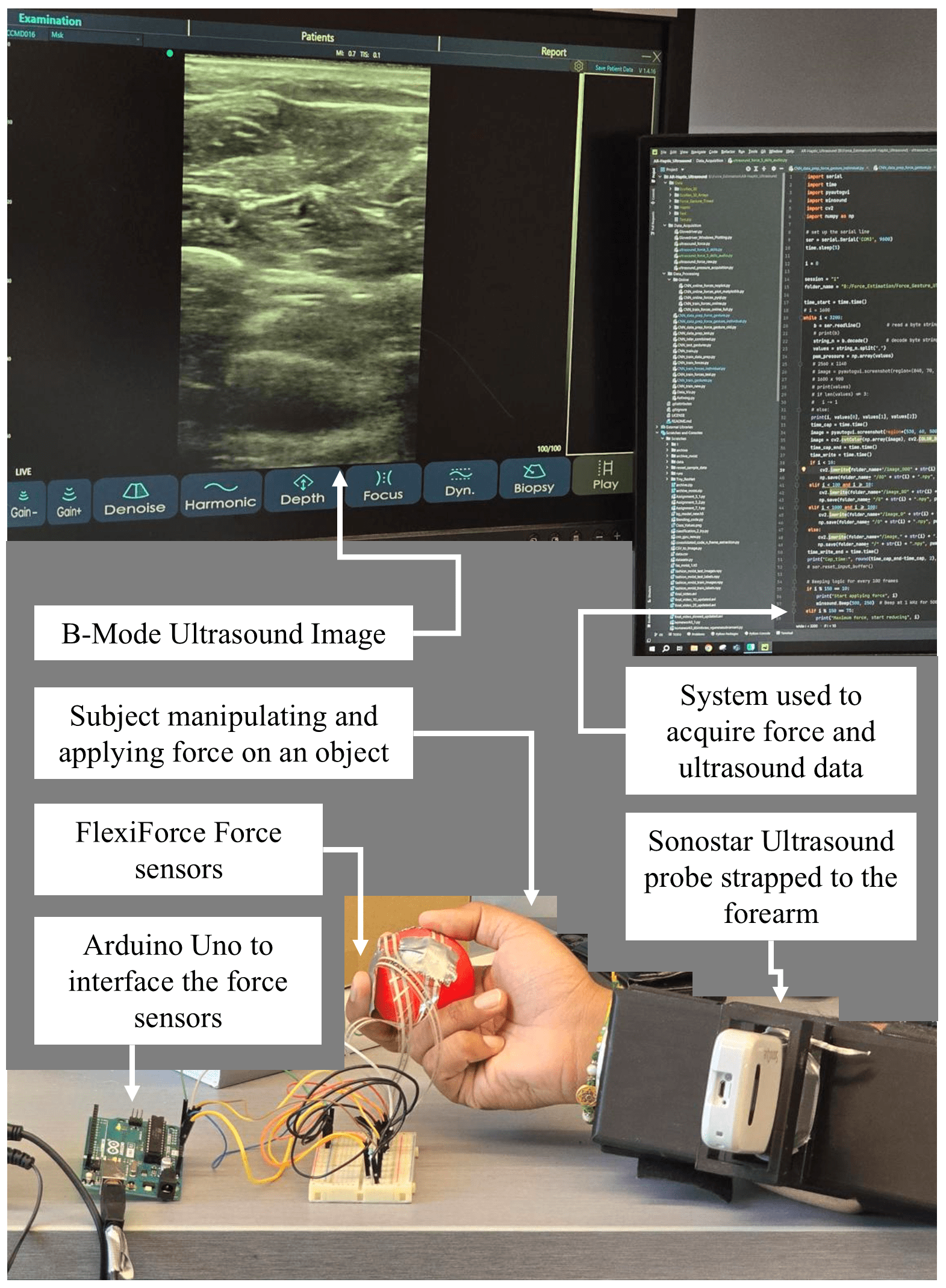}
        \caption{}
    \end{subfigure}
    \hfill
    \begin{subfigure}[t]{0.47\textwidth}
        \centering
        \includegraphics[width=\textwidth]{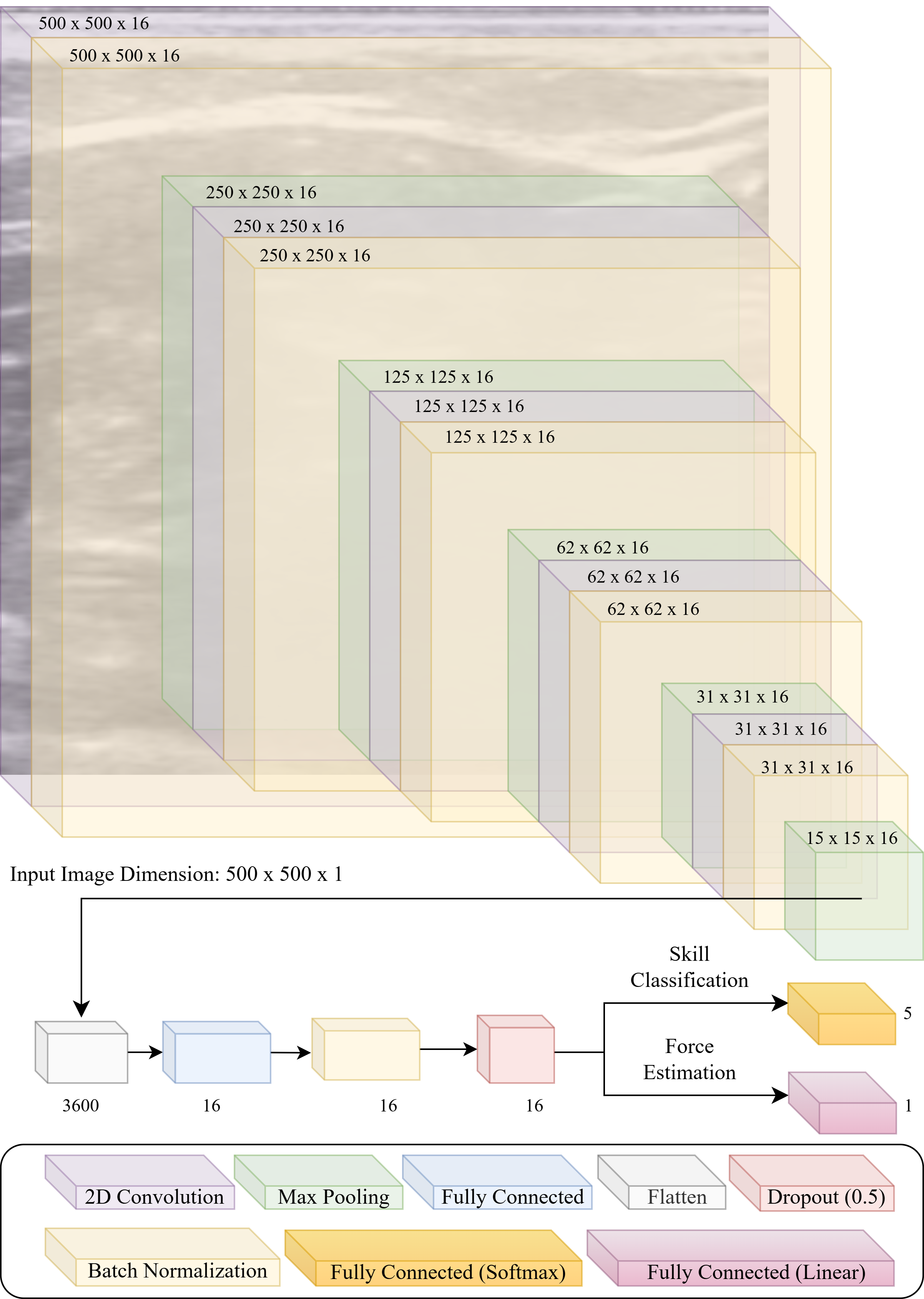}
        \caption{}
    \end{subfigure}
    \caption{(a) Hardware setup for data acquisition: System for acquiring ultrasound and force data. B-mode ultrasound data acquired using a Sonostar probe and 3 FlexiForce sensors used for the thumb, index and middle fingers. An Arduino Uno is used to interface the force sensors with the system, and the data from the probe is transmitted over Wi-Fi. (b) Convolutional Neural Network (CNN) architecture for skill classification and force estimation: The CNN architecture is the same for both tasks, except for the fully connected layer with softmax activation and 5 parameters. For the force estimation task, the fully connected layer has a linear activation and 1 output parameter.}
    \label{fig:system_and_architecture}
\end{figure}

Both the classification and continuous force estimation tasks use a common convolutional neural network (CNN) backbone to extract spatial features from \(500 \times 500 \times 1\) ultrasound image (\(I\)) as shown in \ref{fig:system_and_architecture}(b). The model is adapted from \cite{bimbraw2023simultaneous} and consists of five convolutional layers. Each layer applies a convolution operation (\(*\)) with a \(3 \times 3\) kernel, followed by Rectified Linear Unit activation (\(relu\)) and batch normalization (\(bn\)), as defined in equation \ref{eq:conv_layers}.
\begin{equation}
Z_l = I * f_l, \quad A_l = relu(Z_l), \quad B_l = bn(A_l)
\label{eq:conv_layers}
\end{equation}
Max pooling (\(mp\)), described in equation \ref{eq:max_pool} is applied after each layer, progressively reducing the spatial dimensions.
\begin{equation}
M_l = mp(B_l)
\label{eq:max_pool}
\end{equation}
The feature map is flattened into a one-dimensional vector described in equation \ref{eq:flatten}.
\begin{equation}
FL = fl(M_5)
\label{eq:flatten}
\end{equation}

In both models, \(W_1\) represents the weight matrix in the dense layer after flattening, which maps the flattened feature vector (\(FL\)) to the hidden layer (\(D_1\)) described in equation \ref{eq:dense_layer}.
\begin{equation}
D_1 = relu(FL \cdot W_1)
\label{eq:dense_layer}
\end{equation}
where \(W_1\) is a \((m \times n)\) matrix, with \(m\) being the size of the flattened feature vector and \(n\) being the number of units in the dense layer. This is followed by a drop

The dropout layer prevents overfitting by randomly setting a fraction \(p\) of the input units to zero during training described in equation \ref{eq:dropout_equation}.
\begin{equation}
Dr_1 = drop(D_1, p)
\label{eq:dropout_equation}
\end{equation}
where \(p\) is the dropout rate, and \(D_1\) is the input to the dropout layer.

The skill classification model has 67,525 trainable parameters and predicts one of five manipulation skills. The dense layer maps the features from the last dropout layer (\(Dr_1\)) to the 5 output classes using a weight matrix (\(W_{skill}\)) and a softmax activation:
\begin{equation}
y_{skill} = softmax(Dr_1 \cdot W_{skill})
\label{eq:skill_output}
\end{equation}
where \(W_{skill}\) is a \(n \times 5\) weight matrix, with \(n\) being the number of units in \(Dr_1\).

The force estimation model has 67,457 trainable parameters and predicts continuous values. Its final dense layer maps the features from \(Dr_1\) to a single output using a weight matrix (\(W_{force}\)) and a linear activation:
\begin{equation}
y_{force} = Dr_1 \cdot W_{force}
\label{eq:force_output}
\end{equation}
where \(W_{force}\) is a \(n \times 1\) weight matrix.

\subsection{Analysis}

Forearm ultrasound data was used to estimate both manipulation skills and applied force levels. A convolutional neural network (CNN) architecture based on \cite{bimbraw2023simultaneous} was used for both tasks. Each experiment was repeated three times, and results are reported as the mean and standard deviation across 5-fold cross-validation.

\subsubsection{Skill Classification}

For skill classification, each subject's dataset included 2000 ultrasound frames with synchronized force measurements, of which 20\% was reserved for testing. Training and evaluation were performed across 5 folds, resulting in a total of \(7 \text{ subjects} \times 3 \text{ iterations per fold} \times 5 \text{ folds} = 105 \text{ experiments}\). The model was trained for 10 epochs using the Adam optimizer (\(\texttt{lr}=1 \times 10^{-4}\)), minimizing sparse categorical cross-entropy. Performance was evaluated using classification accuracy.

\subsubsection{Force Estimation}

For force estimation, models were trained on data for each subject and manipulation skill, with 20\% reserved for testing. This resulted in \(7 \text{ subjects} \times 3 \text{ iterations per fold} \times 5 \text{ folds} \times 5 \text{ skills per subject} = 525 \text{ experiments}\). The regression model, sharing the same CNN backbone as the classifier, predicts a single continuous force value. The model was trained for 20 epochs using the Adam optimizer (\(\texttt{lr}=1 \times 10^{-3}\)), minimizing mean absolute error (MAE) and evaluated using root mean squared error (RMSE).

\subsubsection{Interpretability Analysis}
Grad-CAM (Gradient-weighted Class Activation Mapping) was applied to visualize regions of the ultrasound images that the CNN focused on for continuous force estimation tasks. For each subject, Grad-CAM heatmaps were generated by back-propagating gradients through the final convolutional layer of the CNN \cite{selvaraju2017grad}. Grad-CAM is used to visualize which regions of the input ultrasound image contributed the most to the prediction of a particular manipulation skill. Let \( y_{\text{force}} \) denote the model's output for continuous force prediction. The gradients of \( y_{\text{force}} \) with respect to the feature maps of the final convolutional layer \( A^k \) are then computed to obtain \(\frac{\partial y_{\text{force}}}{\partial A_{ij}^k}\), where \( A_{ij}^k \) represents the activation of the \( k \)-th feature map at position \((i, j)\). These gradients are averaged spatially to compute importance weights for each feature map, defined in Equation \ref{eq:alpha_k}.

\begin{equation}
    \alpha_k = \frac{1}{Z} \sum_{i} \sum_{j} \frac{\partial y_{\text{force}}}{\partial A_{ij}^k}
    \label{eq:alpha_k}
\end{equation}

where \( Z \) is the total number of pixels in the feature map. The importance weights \( \alpha_k \) are used to compute a weighted combination of the feature maps leading to the heatmap defined in Equation \ref{eq:heatmap}.

\begin{equation}
L_{\text{Grad-CAM}}^{\text{force}} = \text{ReLU}\left(\sum_k \alpha_k A^k\right)
\label{eq:heatmap}
\end{equation}

The resulting heatmap \(L_{\text{Grad-CAM}}^{\text{force}} \) highlights the regions in the input image that were most critical for predicting the continuous force application relevant to the manipulation skill.

To enhance interpretabiility, a dynamic weight allocation mechanism was applied to prioritize convolutional layers based on their importance to specific force patterns. Earlier layers are more effective at identifying low-level spatial details such as textures or edges, while deeper layers excel at recognizing high-level patterns. By dynamically assigning weights to these layers, the method emphasizes features that contribute to more precise and meaningful heatmaps, avoiding overly vague visualizations that hinder interpretability. The layer weights are adjusted based on the identified skill class, with high importance layers receiving a weight of 1.0, medium importance layers receiving a weight of 0.75, low importance layers weighted at 0.35 and very low importance layers receiving a weight of 0.15.To further enhance interpretability, guided backpropagation is applied. Guided backpropagation modifies the gradient computation by restricting the flow of negative gradients, ensuring that only positive contributions to the prediction are propagated:

\begin{equation}
    g(x) = \text{ReLU}\left(\text{ReLU}(x) \cdot \text{ReLU}\left(\frac{\partial L}{\partial x}\right)\right)
\end{equation}

\begin{figure}
    \centering
    \includegraphics[width=\textwidth]{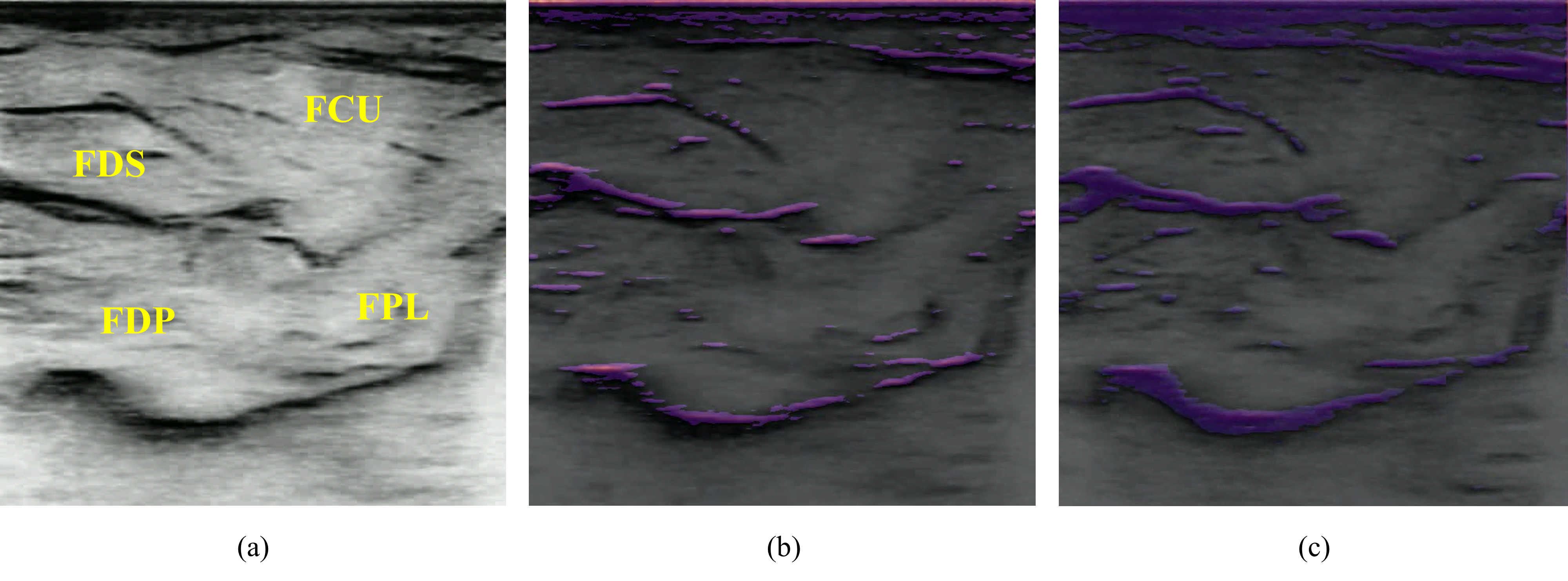}
    \caption{Different muscles and their activation: (a) Flexor digitorum superficialis (FDS), flexor carpi ulnaris (FCU), flexor digitorum profundus (FDP) and flexor pollicis longus (FPL). The changes in the muscle activations visualized using Grad-CAM for manipulation skill 2 in (b) and (c). The bright areas correspond to the regions of the image contribute to the CNN's continuous force estimation.}
    \label{fig:muscles}
\end{figure}

The regions in the forearm ultrasound data can be visualized in Figure \ref{fig:muscles}. A physician was asked to view the videos of Grad-CAM based muscle activations for each subject and skill. They were then asked to describe what muscle groups are activated. The physician answered FDS, FDP, FPL, FCU or too much artifact. These results were aggregated to understand how the muscles perform different manipulation skills and forces.

\hspace{0.02\textwidth}%

\subsection{Quantification Metrics}

Classification accuracy (Acc) was used to evaluate the model’s performance in identifying manipulation skills, while the Root Mean Squared Error (RMSE) quantified the regression performance in estimating applied force levels. For both Acc and RMSE, the arithmetic mean ($\mu$) was calculated to summarize the metrics across subjects and tasks.

\subsubsection{Classification Accuracy}
The accuracy percentage was calculated to evaluate the skill classification model’s performance in predicting manipulation skills accurately, defined in equation \ref{acc_eqn}.

\begin{equation}
    \text{Acc} = \frac{TP + TN}{N} \times 100
    \label{acc_eqn}
\end{equation}

where $TP$ represents the number of True Positives, $TN$ the number of True Negatives, and $N$ the total sample size. $TP$ is the outcome where the model correctly predicts the positive class, while $TN$ is the outcome where the model correctly predicts the negative class. The classification error percentage was calculated by subtracting Acc from 100.

\subsubsection{Root Mean Squared Error (RMSE)}
RMSE was used to evaluate the force estimation model's performance, defined in equation \ref{rmse}.

\begin{equation}
    \text{RMSE} = \sqrt{\frac{1}{N} \sum_{i=1}^{N} (y_i - \hat{y}_i)^2}
    \label{rmse}
\end{equation}

where $N$ is the total sample size, $y_i$ is the true force measurement for sample $i$, and $\hat{y}_i$ is the predicted force value for sample $i$.

\subsubsection{Arithmetic Mean ($\mu$)}
To provide a summary of both Acc and RMSE metrics across subjects and tasks, the arithmetic mean defined in equation \ref{eq:mu} was calculated.

\begin{equation}
    \mu = \frac{1}{N} \sum_{i=1}^{N} a_i
    \label{eq:mu}
\end{equation}

where $N$ represents the number of values, and $a_i$ is the individual value from a set of Acc or RMSE values.

\subsubsection{Standard Deviation ($\sigma$)}
The standard deviation ($\sigma$) defined in equation \ref{eq:sigma} was calculated to measure the dispersion of the Acc and RMSE values.

\begin{equation}
    \sigma = \sqrt{\frac{1}{N} \sum_{i=1}^{N} (a_i - \mu)^2}
    \label{eq:sigma}
\end{equation}

where $N$ is the number of values, $a_i$ is the individual value, and $\mu$ is the arithmetic mean of the dataset.

\subsection{Experimental Design}
The experimental setup was designed to evaluate the proposed framework's performance for classification of manipulation skills and continuous estimation of applied force. 
\subsubsection{System Configuration}
The experiments were conducted on a computing system equipped with an NVIDIA GeForce RTX 2070 SUPER GPU, an AMD Ryzen 7 2700X Eight-Core Processor, and 31.91 GB of RAM. Python 3.7 was used for implementing the data processing and model training workflow. TensorFlow 1.14.0 was used to train and evaluate the model performance \cite{tensorflow2015-whitepaper}. 
\subsubsection{Data Preprocessing}
Ultrasound images were loaded as grayscale `.png` files, normalized, and reshaped into 4D tensors of size \((\text{samples}, \text{height}, \text{width}, 1)\).

For skill classification, the labels were the numeric values associated with each manipulation skill. Per subject, the dataset for each manipulation skill contained 2000 image-label pairs, which were split into the first 1600 samples per skill as the training set, totaling 8000 samples across the five skills. The remaining 400 samples per skill, totaling 2000 samples were used as the test set.
The data was balanced across all five manipulation skills to prevent class imbalance. Processed datasets were saved as \texttt{.npy} files.

For the continuous force estimation, preprocessing involved preparing ultrasound images and corresponding force values to predict continuous force outputs. Force ground truth labels were computed as the average of three sensor readings. Per subject and skill, the dataset for each manipulation skill contained 2000 image-label pairs, wherein the first 1600 samples were used for training and the remaining 400 samples were used for testing.

\section{Results}\label{sec:results}
In this study, the CNN model’s performance in classifying manipulation skills and estimating force was assessed across seven subjects. Additionally, Grad-CAM was used to visualize the regions of ultrasound images that contributed most to the model’s predictions, providing interpretability for both classification and regression tasks. \
\subsection{Skill Classification}
We hypothesized that the CNN model can effectively classify manipulation skills using forearm ultrasound data, achieving high accuracy across subjects and folds. Averaged over three iterations, the classification model achieved a five fold cross validation test accuracy of 94.9 $\pm$ 10.2 \% across the seven subjects, demonstrating strong performance in distinguishing between manipulation skills using forearm ultrasound data. The training accuracy was 100.0 $\pm$ 0.1\% accuracy on the training set, exceeding 99 \% for every subject. 
\begin{figure}[ht]
    \centering
    % Top row: Fold-wise results
    \begin{subfigure}{0.48\textwidth}
        \centering
        \includegraphics[width=\textwidth]{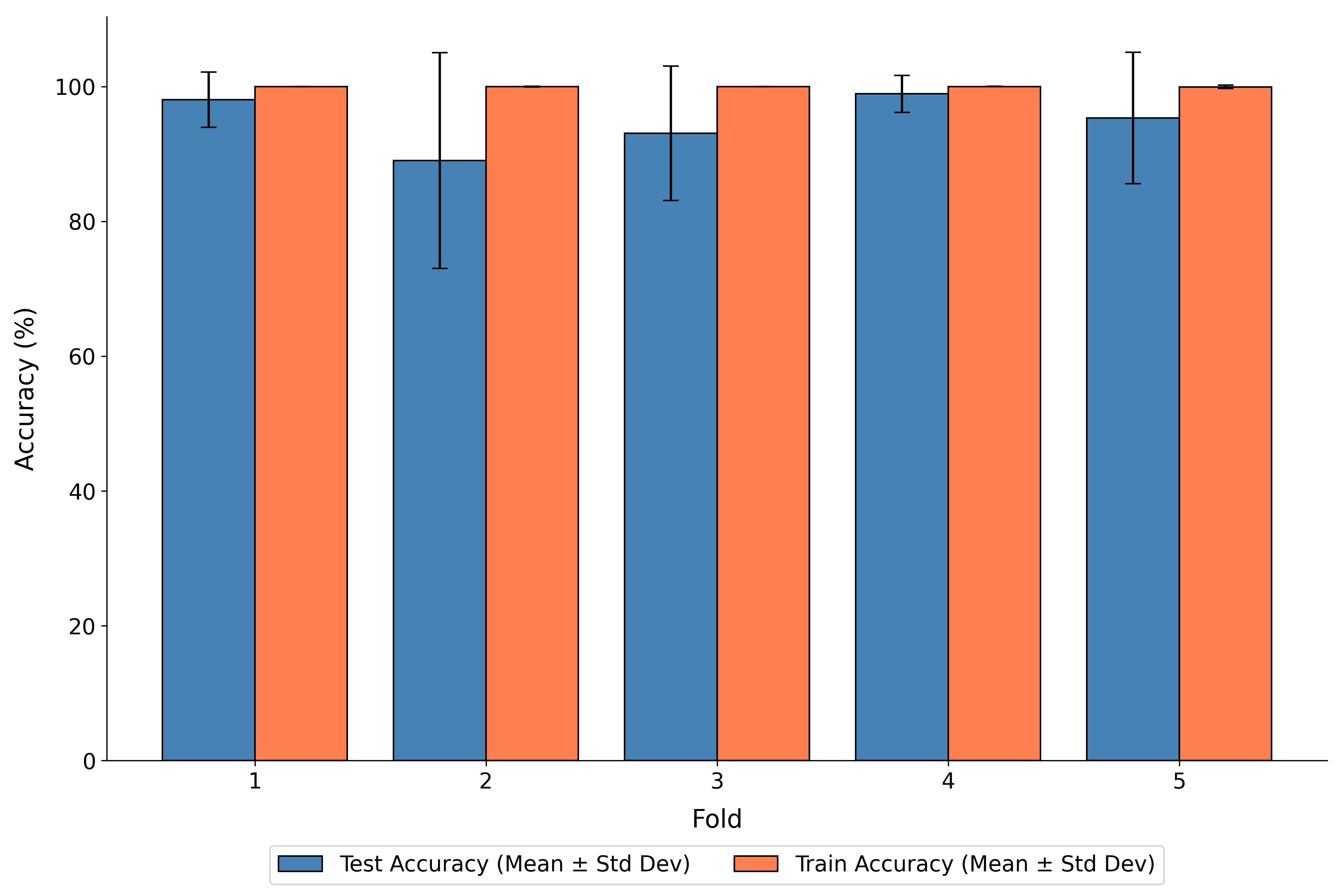}
        \caption{Fold-wise classification results}
        \label{fig_fold_accuracy_stats}
    \end{subfigure}\hspace{0.02\textwidth}%
    % Top row: Subject-wise results
    \begin{subfigure}{0.48\textwidth}
        \centering
        \includegraphics[width=\textwidth]{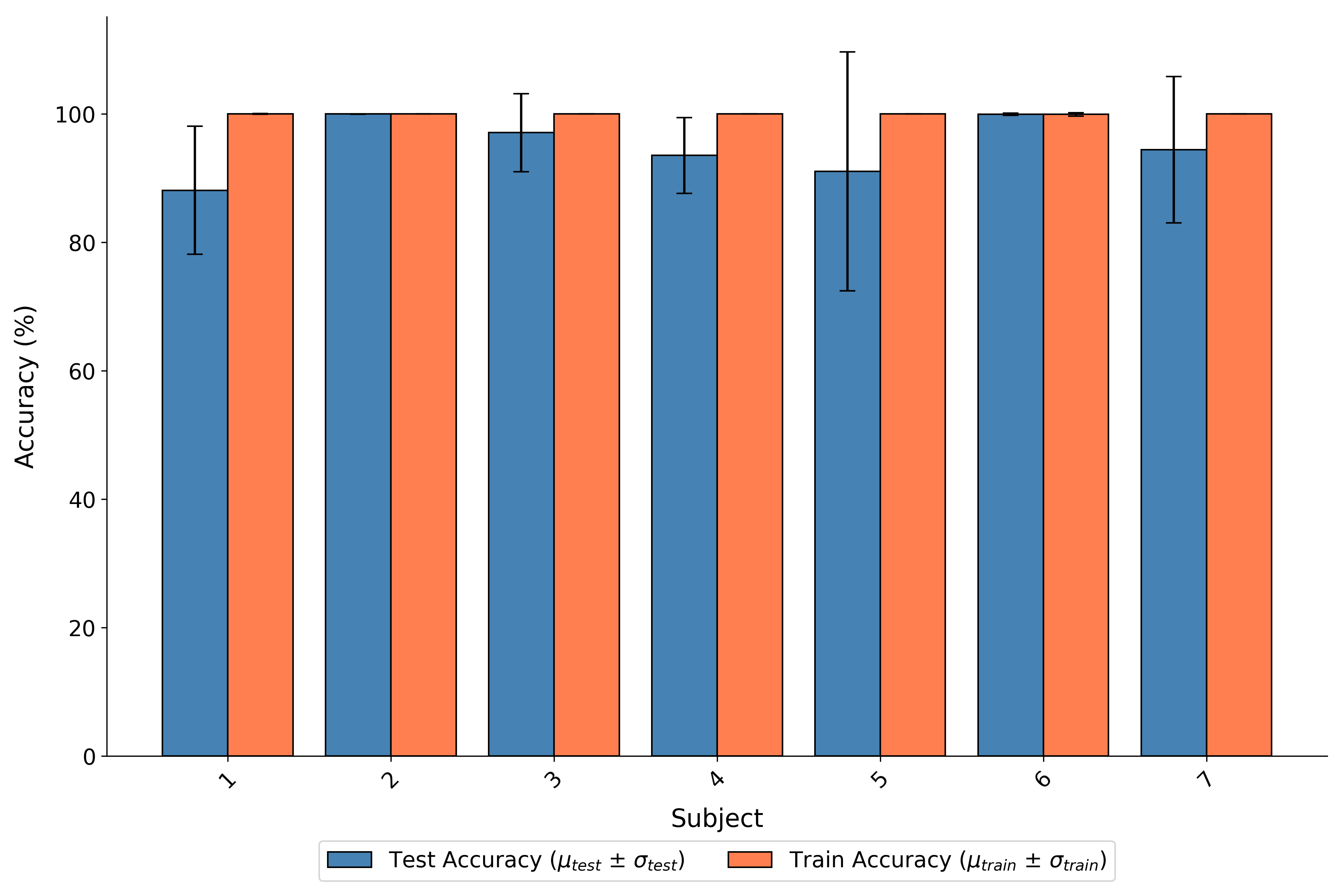}
        \caption{Subject-wise classification results}
        \label{fig:subject-classification}
    \end{subfigure}
    \caption{Comparison of fold-wise and subject-wise classification results.}
    \label{fig:combined_classification}
\end{figure}

\subsubsection{Fold-Wise Classification Results}
% \begin{figure}[ht]
%     \centering
%     \includegraphics[width=0.75\textwidth]{figures/fold_classification_plot.png}
%     \caption{Fold-wise classification results}
%     \label{fig_fold_accuracy_stats}
% \end{figure}

\begin{table}[h]
    \centering
    \caption{Fold-Wise Classification Results}
    \begin{tabular}{|c|c|c|c|c|}
        \hline
        \textbf{Fold} & $\mu_{\text{train}}$ & $\sigma_{\text{train}}$ & $\mu_{\text{test}}$ & $\sigma_{\text{test}}$ \\ \hline
        1 & 100.00 & 0.00 & 98.04 & 4.11 \\ \hline
        2 & 99.99 & 0.04 & 89.02 & 15.98 \\ \hline
        3 & 100.00 & 0.00 & 93.06 & 9.97 \\ \hline
        4 & 99.99 & 0.04 & 98.91 & 2.74 \\ \hline
        5 & 99.95 & 0.23 & 95.33 & 9.75 \\ \hline
    \end{tabular}
    \label{table_fold_accuracy_stats}
\end{table}

We hypothesize that the test accuracy of the CNN model will vary across folds, with some folds achieving higher accuracy and lower variability, indicating better generalization to unseen data. The fold-wise classification results are shown in Figure \ref{fig_fold_accuracy_stats} and Table \ref{table_fold_accuracy_stats}. Across all folds, the mean training accuracy ($\mu_{\text{train}}$) remains nearly perfect (99.95\% to 100.00\%) with negligible variability, as indicated by $\sigma_{\text{train}}$ (maximum 0.23 in Fold 5). The mean test accuracy ($\mu_{\text{test}}$) was the highest for fold 4 achieves $\mu_{\text{test}}$ at 98.91\% with the lowest variability ($\sigma_{\text{test}} = 2.74$), and the lowest for fold 2 with the poorest $\mu_{\text{test}}$ at 89.02\% with the highest variability ($\sigma_{\text{test}} = 15.98$). These results support the hypothesis that the model's test accuracy and variability are influenced by differences across folds, with some folds demonstrating better generalization performance.

\subsubsection{Subject-Wise Classification Results}
We hypothesize that the test accuracy of the CNN model will vary across subjects, with some subjects exhibiting higher accuracy and lower variability. This is because of subject-specific differences in ultrasound data quality and manipulation skills. A summary of train and test accuracies across subjects highlights the model’s stability in identifying manipulation skills can be found in Figure \ref{fig:subject-classification} and Table \ref{subject-classification}. 
The mean training accuracy ($\mu_{\text{train}}$) is high across all subjects, ranging from 99.92\% (Subject 6) to 100.00\% (Subjects 2, 3, 4, 5, and 7). 
The standard deviation in training accuracy ($\sigma_{\text{train}}$) is minimal, with a maximum of 0.27\% (Subject 6), indicating consistent training performance for all subjects.
The mean test accuracy ($\mu_{\text{test}}$) is variable across subjects, with subject 2 achieving the best $\mu_{\text{test}}$ at 99.99\% with negligible variability ($\sigma_{\text{test}} = 0.04$) and subject 1 achieving the lowest $\mu_{\text{test}}$ at 88.10\%, with higher variability ($\sigma_{\text{test}} = 9.96$).
Other subjects show intermediate performance, with $\mu_{\text{test}}$ ranging from 91.04\% (Subject 5) to 99.95\% (Subject 6).
Subject 5 exhibits the highest $\sigma_{\text{test}}$ (18.59\%), suggesting more inconsistent performance in classification compared to other subjects.
Subject 7 also has elevated $\sigma_{\text{test}}$ (11.37\%), though with a higher mean test accuracy of 94.42\%.
Subjects 2, 3, and 6 exhibit both high $\mu_{\text{test}}$ and low $\sigma_{\text{test}}$ (e.g., Subject 2: 99.99\% $\pm$ 0.04\%), indicating reliable generalization for these subjects.
These findings confirm the hypothesis that test accuracy and variability are subject-dependent, influenced by individual differences in ultrasound data and manipulation skill patterns.

\begin{table}
\centering
\caption{Subject-wise Classification Results}
\label{tab:subject_classification_results}
\begin{tabular}{|c|c|c|c|c|}
\hline
Subject & $\mu_{train}$ & $\sigma_{train}$ & $\mu_{test}$ & $\sigma_{test}$ \\\hline
   1 &  99.99 &    0.05 &                88.10 &                    9.96 \\\hline
   2 & 100.00 &    0.00 &                99.99 &                    0.04 \\\hline
   3 & 100.00 &    0.00 &                97.08 &                    6.06 \\\hline
   4 & 100.00 &    0.00 &                93.52 &                    5.91 \\\hline
   5 & 100.00 &    0.00 &                91.04 &                   18.59 \\\hline
   6 & 99.92 &     0.27 &                99.95 &                    0.18 \\\hline
   7 & 100.00 &    0.00 &                94.42 &                   11.37 \\\hline
\end{tabular}
\label{subject-classification}
\end{table}

\subsection{Force Estimation}
The five fold cross validation force estimation performance, averaged over the folds and three iterations yielded an average test RMSE of 0.51 $\pm$ 0.19 N across all skills and subjects. 
The train RMSE was obtained to be 0.37 $\pm$ 0.14 N. 
\begin{figure}[ht]
    \centering
    \begin{subfigure}{0.48\textwidth}
        \centering
        \includegraphics[width=\textwidth]{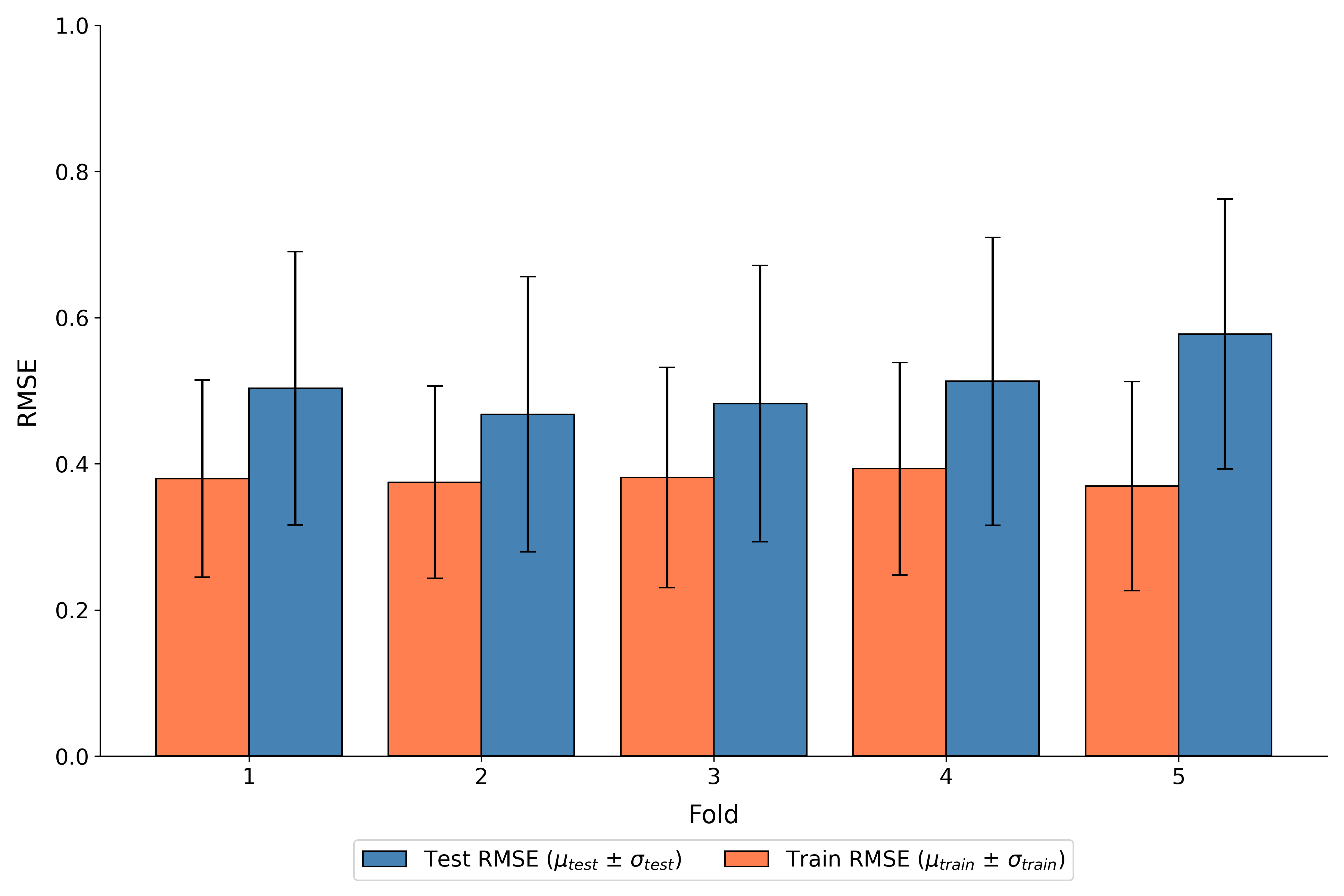}
        \caption{Fold-wise RMSE results}
        \label{fig:fold-regression}
    \end{subfigure}\hspace{0.02\textwidth}
    \begin{subfigure}{0.48\textwidth}
        \centering
        \includegraphics[width=\textwidth]{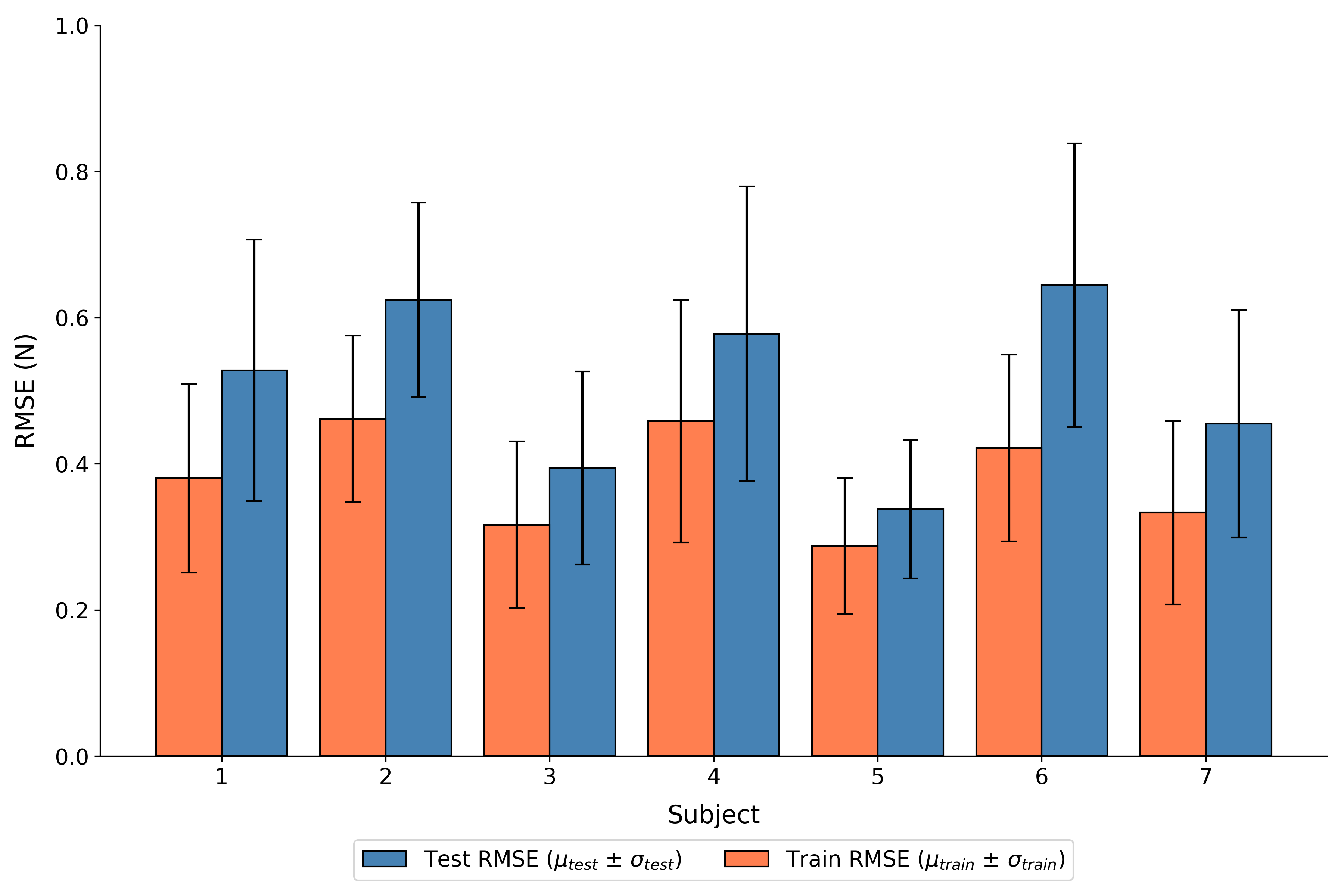}
        \caption{Subject-wise RMSE results}
        \label{fig:subject-regression}
    \end{subfigure}
    \caption{Comparison of fold-wise and subject-wise RMSE results.}
    \label{fig:combined_regression}
\end{figure}
\subsubsection{Fold-Wise Force Estimation}

\begin{table}[h]
    \centering
    \caption{Fold-Wise RMSE Results ($\mu$ and $\sigma$) with Percentages of Maximum Sensor Range (4 N)}
    \begin{tabular}{|c|c|c|c|c|c|c|c|c|}
        \hline
        \textbf{Fold} & $\mu_{\text{train}}$ & $\sigma_{\text{train}}$ & $\mu_{\text{test}}$ & $\sigma_{\text{test}}$ & $\mu_{\text{train}}$ (\%) & $\sigma_{\text{train}}$ (\%) & $\mu_{\text{test}}$ (\%) & $\sigma_{\text{test}}$ (\%) \\ \hline
        1 & 0.38 & 0.13 & 0.50 & 0.19 & 9.49 & 3.37 & 12.58 & 4.67 \\ \hline
        2 & 0.37 & 0.13 & 0.47 & 0.19 & 9.37 & 3.29 & 11.69 & 4.70 \\ \hline
        3 & 0.38 & 0.15 & 0.48 & 0.19 & 9.53 & 3.77 & 12.06 & 4.73 \\ \hline
        4 & 0.39 & 0.15 & 0.51 & 0.20 & 9.83 & 3.63 & 12.82 & 4.92 \\ \hline
        5 & 0.37 & 0.14 & 0.58 & 0.18 & 9.24 & 3.58 & 14.44 & 4.61 \\ \hline

    \end{tabular}
    \label{table_fold_rmse_percent}
\end{table}
We hypothesize that the test RMSE will exhibit moderate variability across folds, reflecting differences in generalization performance for each fold. The average RMSE for each fold is shown in Figure \ref{fig:fold-regression} and Table \ref{table_fold_rmse_percent}. 
The mean training RMSE across folds is consistently low, ranging from 0.37 to 0.39 N with he percentage of training RMSE relative to the maximum sensor range (4 N) ranges from 9.24\% to 9.83\%. 
The standard deviation of training RMSE ($\sigma_{\text{train}}$) has values between 0.13 and 0.15 N (3.29\% to 3.77\% of 4 N).
The mean testing RMSE shows moderate variability across folds, with the lowest for fold 2 with $\mu_{\text{test}}$ of 0.47 N (11.69\% of 4 N) and the highest for fold 5 with $\mu_{\text{test}}$ of 0.58 N (14.44\% of 4 N).
The standard deviation of testing RMSE ($\sigma_{\text{test}}$) is higher than training, ranging from 0.18 to 0.20 N (4.61\% to 4.92\% of 4 N).
These results confirm the hypothesis that test RMSE shows fold-dependent variability, with some folds demonstrating higher consistency and lower errors.

\subsubsection{Subject-Wise Force Estimation}

\begin{table}[h]
    \centering
    \caption{Subject-Wise Regression RMSE Results (Mean and Standard Deviation) with Percentages of Maximum (4 N)}
    \begin{tabular}{|c|c|c|c|c|c|c|c|c|}
        \hline
        \textbf{Subject} & $\mu_{\text{train}}$ & $\sigma_{\text{train}}$ & $\mu_{\text{test}}$ & $\sigma_{\text{test}}$ & $\mu_{\text{train}}$ (\%) & $\sigma_{\text{train}}$ (\%) & $\mu_{\text{test}}$ (\%) & $\sigma_{\text{test}}$ (\%) \\ \hline
        1 & 0.38 & 0.13 & 0.53 & 0.18 & 9.50 & 3.23 & 13.20 & 4.47 \\ \hline
        2 & 0.46 & 0.11 & 0.62 & 0.13 & 11.54 & 2.85 & 15.61 & 3.32 \\ \hline
        3 & 0.32 & 0.11 & 0.39 & 0.13 & 7.91 & 2.85 & 9.85 & 3.30 \\ \hline
        4 & 0.46 & 0.17 & 0.58 & 0.20 & 11.46 & 4.15 & 14.45 & 5.04 \\ \hline
        5 & 0.29 & 0.09 & 0.34 & 0.09 & 7.18 & 2.32 & 8.45 & 2.36 \\ \hline
        6 & 0.42 & 0.13 & 0.64 & 0.19 & 10.54 & 3.19 & 16.11 & 4.85 \\ \hline
        7 & 0.33 & 0.13 & 0.45 & 0.16 & 8.33 & 3.13 & 11.37 & 3.89 \\ \hline

    \end{tabular}
    \label{table_subject_rmse_percent}
\end{table}

We hypothesize that test RMSE will vary across subjects, with some subjects exhibiting higher RMSE and greater variability due to individual differences in ultrasound data and skill performance.
The average RMSE for each subject is shown in Figure \ref{fig:subject-regression} and Table \ref{table_subject_rmse_percent}. 
Mean training RMSE ranges from 0.29 N (Subject 5) to 0.46 N (Subjects 2 and 4).
Training RMSE is low across subjects, with percentages of the maximum (4 N) ranging from 7.18\% to 11.54\%.
The standard deviation of training RMSE ($\sigma_{\text{train}}$) is minimal, ranging from 0.09 N (2.32\% of 4 N) to 0.17 N (4.15\% of 4 N), reflecting consistent performance across iterations and folds for training data.
Mean testing RMSE is higher and shows more variability compared to training with subject 5 achieving the lowest $\mu_{\text{test}}$ at 0.34 N (8.45\% of 4 N). 
The highest test RMSE was obtained for subject 6 with $\mu_{\text{test}}$ at 0.64 N (16.11\% of 4 N).
The standard deviation of testing RMSE ($\sigma_{\text{test}}$) varies across subjects, with values between 0.09 N (2.36\% of 4 N) and 0.20 N (5.04\% of 4 N), indicating greater variability in test performance for certain subjects.
Subjects 4 and 6 exhibit higher $\sigma_{\text{test}}$ (5.04\% and 4.85\% of 4 N), suggesting less consistent performance across iterations for these individuals.
Subject 5 demonstrates both the lowest $\mu_{\text{test}}$ and $\sigma_{\text{test}}$, indicating reliable generalization for this subject’s data.
These findings support the hypothesis that subject-dependent factors influence test RMSE, with some subjects showing higher errors and variability.

\subsubsection{Skill-Wise Force Estimation}
\begin{figure}[ht]
    \centering
    \includegraphics[width=0.75\textwidth]{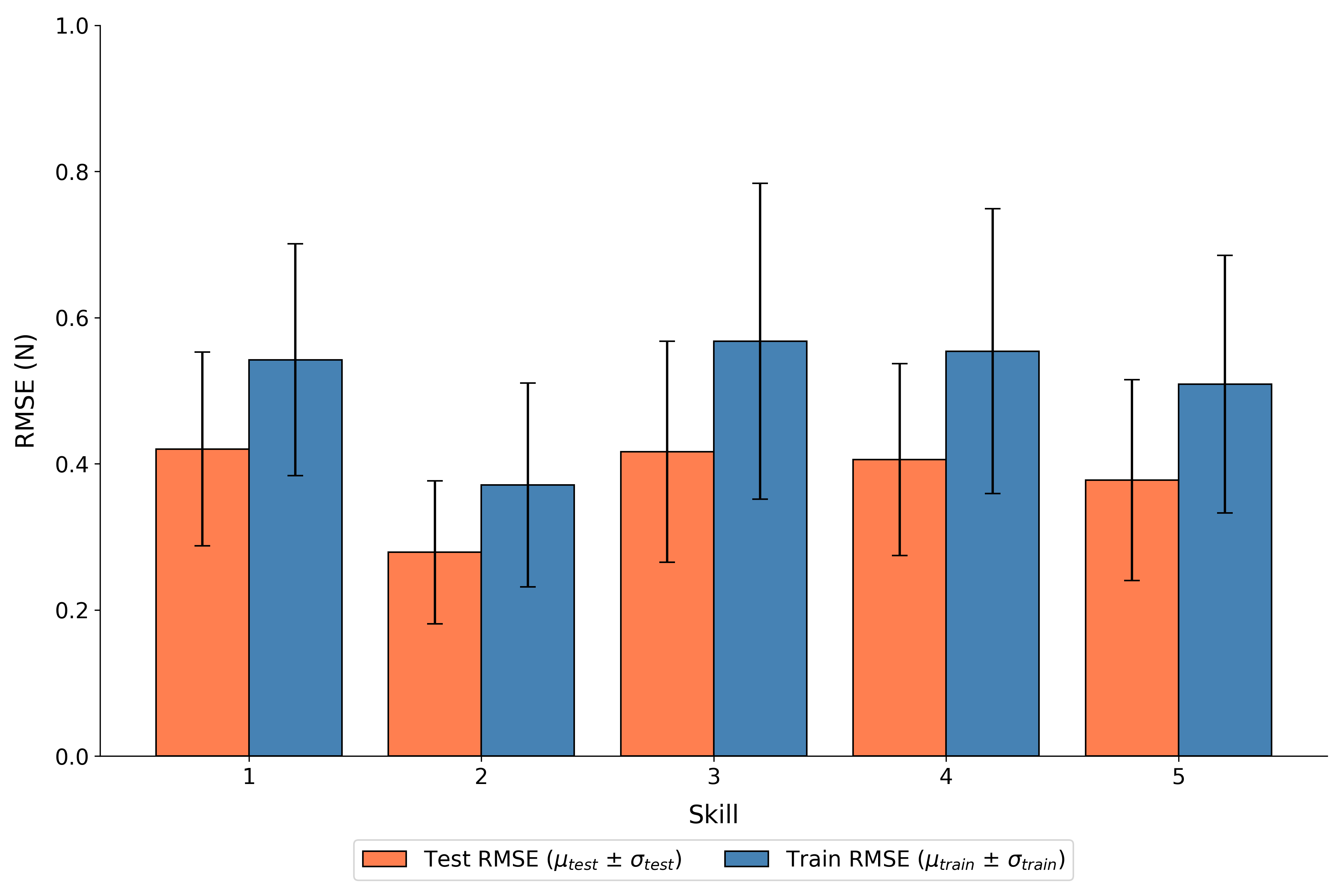}
    \caption{Skill-wise RMSE results}
    \label{fig:skill-regression}
\end{figure}

We hypothesize that test RMSE will vary across skills, with some skills demonstrating higher errors and variability due to differences in task complexity and execution.
The skill-wise force estimation results for both training and testing datasets are summarized in Table \ref{table_class_rmse_percent} and Figure \ref{fig:skill-regression}. 
The mean training RMSE ranges from 0.38 N (Skill 2) to 0.54 N (Skill 3), corresponding to 9.50\% to 13.50\% of the 4 N maximum. 
The standard deviation of training RMSE ($\sigma_{\text{train}}$) is low across all skills, ranging from 0.10 N (2.50\% of 4 N, Skill 5) to 0.16 N (4.00\% of 4 N, Skill 3), reflecting consistent performance during training.
The mean testing RMSE is higher than the training RMSE, ranging from 0.46 N (11.50\% of 4 N, Skill 2) to 0.68 N (17.00\% of 4 N, Skill 3). 
The standard deviation of testing RMSE ($\sigma_{\text{test}}$) shows greater variability compared to training, with values ranging from 0.12 N (3.00\% of 4 N, Skill 2) to 0.22 N (5.50\% of 4 N, Skill 3).

Skill 3 demonstrates the highest RMSE values, with the largest percentage of the 4 N maximum indicating a higher level of variability in data for this skill.
Skill 2 consistently shows the lowest RMSE values, indicating easier generalization and minimal variability.
The variability (standard deviation) in testing is generally higher than in training, suggesting greater heterogeneity in the testing data.
These results validate the hypothesis that task complexity impacts test RMSE, with specific skills showing consistently higher errors and variability.

\begin{table}[h]
    \centering
    \caption{Skill-Wise RMSE Results with Percentages of Maximum (4 N)}
    \begin{tabular}{|c|c|c|c|c|c|c|c|c|}
        \hline
        \textbf{Skill} & $\mu_{\text{train}}$ & $\sigma_{\text{train}}$ & $\mu_{\text{test}}$ & $\sigma_{\text{test}}$ & $\mu_{\text{train}}$ (\%) & $\sigma_{\text{train}}$ (\%) & $\mu_{\text{test}}$ (\%) & $\sigma_{\text{test}}$ (\%) \\ \hline
        1 & 0.49 & 0.14 & 0.61 & 0.19 & 12.25 & 3.50 & 15.25 & 4.75 \\ \hline
        2 & 0.38 & 0.13 & 0.46 & 0.12 & 9.50 & 3.25 & 11.50 & 3.00 \\ \hline
        3 & 0.54 & 0.16 & 0.68 & 0.22 & 13.50 & 4.00 & 17.00 & 5.50 \\ \hline
        4 & 0.45 & 0.12 & 0.59 & 0.15 & 11.25 & 3.00 & 14.75 & 3.75 \\ \hline
        5 & 0.40 & 0.10 & 0.52 & 0.14 & 10.00 & 2.50 & 13.00 & 3.50 \\ \hline
    \end{tabular}
    \label{table_class_rmse_percent}
\end{table}

\subsection{Interpretability Analysis}
Grad-CAM visualization provided insights into the regions of the ultrasound images that the model focused on for predicting manipulation skills and force levels. Splitscreen videos were generated with an overlay of the heatmap on the original ultrasound frames, which were utilized to identify key muscle groups. 

The Grad-CAM results were analyzed to identify key muscle activations across different manipulation skills (1 through 5). The muscles visible in the images were flexor pollicis longus (FPL), flexor digitorum profundus (FDP), flexor digitorum superficialis (FDS), and flexor carpi ulnaris (FCU), which play distinct roles in hand and wrist movements, and are shown in Figure \ref{fig:muscles}(a). FPL and FDP are responsible for finger flexion, FDS supports finer grasping actions, and FCU contributes to wrist stability and movement. Additionally, the occurrences of too many artifacts were noted. 

\begin{figure}[ht]
    \centering
    \includegraphics[width=0.75\textwidth]{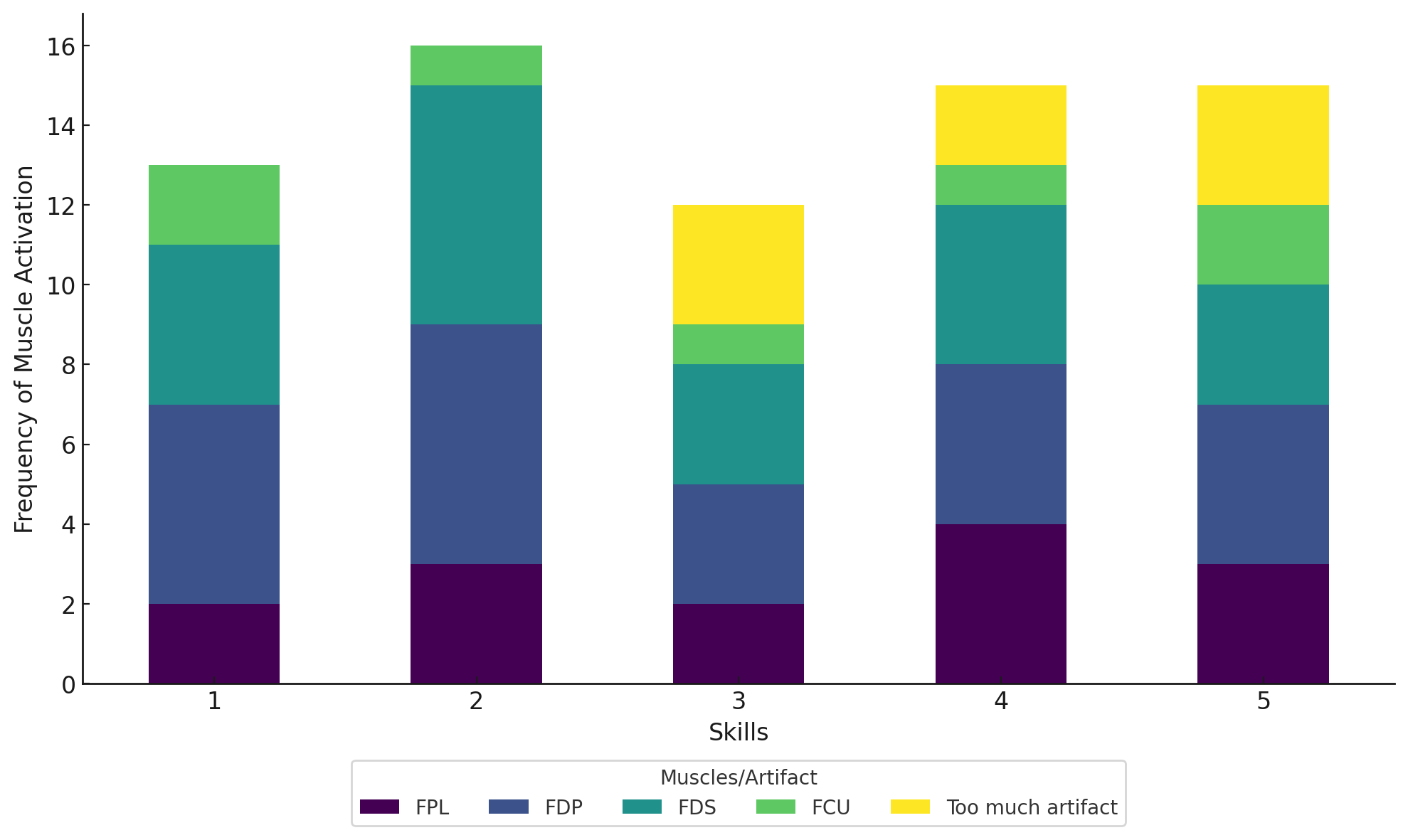}
    \caption{Skill-wise Grad-CAM based muscle activations}
    \label{fig:skill-gradcam}
\end{figure}

We hypothesize that different manipulation skills will activate specific muscle groups consistently, with FDS and FDP being the most frequently activated muscle, and that skills with higher artifacts will exhibit less interpretable results. The summarized findings are presented in Table~\ref{tab:gradcam_results} and Figure \ref{fig:skill-gradcam}.

\begin{table}[h!]
\centering
\caption{Frequency of muscle activations and artifacts across manipulation tasks.}
\label{tab:gradcam_results}
\begin{tabular}{|c|c|c|c|c|c|}
\hline
\textbf{Skill} & \textbf{FPL} & \textbf{FDP} & \textbf{FDS} & \textbf{FCU} & \textbf{Too much artifact} \\
\hline
1 & 2 & 5 & 4 & 2 & 0 \\\hline
2 & 3 & 6 & 6 & 1 & 0 \\\hline
3 & 2 & 3 & 3 & 1 & 3 \\\hline
4 & 4 & 4 & 4 & 1 & 2 \\\hline
5 & 3 & 4 & 3 & 2 & 3 \\
\hline
\end{tabular}
\end{table}

The results show that FDP exhibited the highest frequency of activation across all tasks, particularly for skills 1 and 2. FPL and FDS were also consistently activated across tasks, while FCU showed less frequent involvement. Skills 3 and 5 experienced the highest frequency of artifacts, making their results less interpretable. Skills 1 and 2 had no reported artifacts, resulting in cleaner Grad-CAM outputs. Skills 1 and 2 demonstrated more consistent activations, while skills 3--5 showed greater variability and interference from artifacts. These findings confirm the hypothesis that FDP shows the highest activation across tasks, FCU shows the least activation across tasks, and skills with higher artifacts (e.g., Skills 3 and 5) exhibit reduced interpretability.

\begin{figure}[ht]
    \centering
    \includegraphics[width=0.75\textwidth]{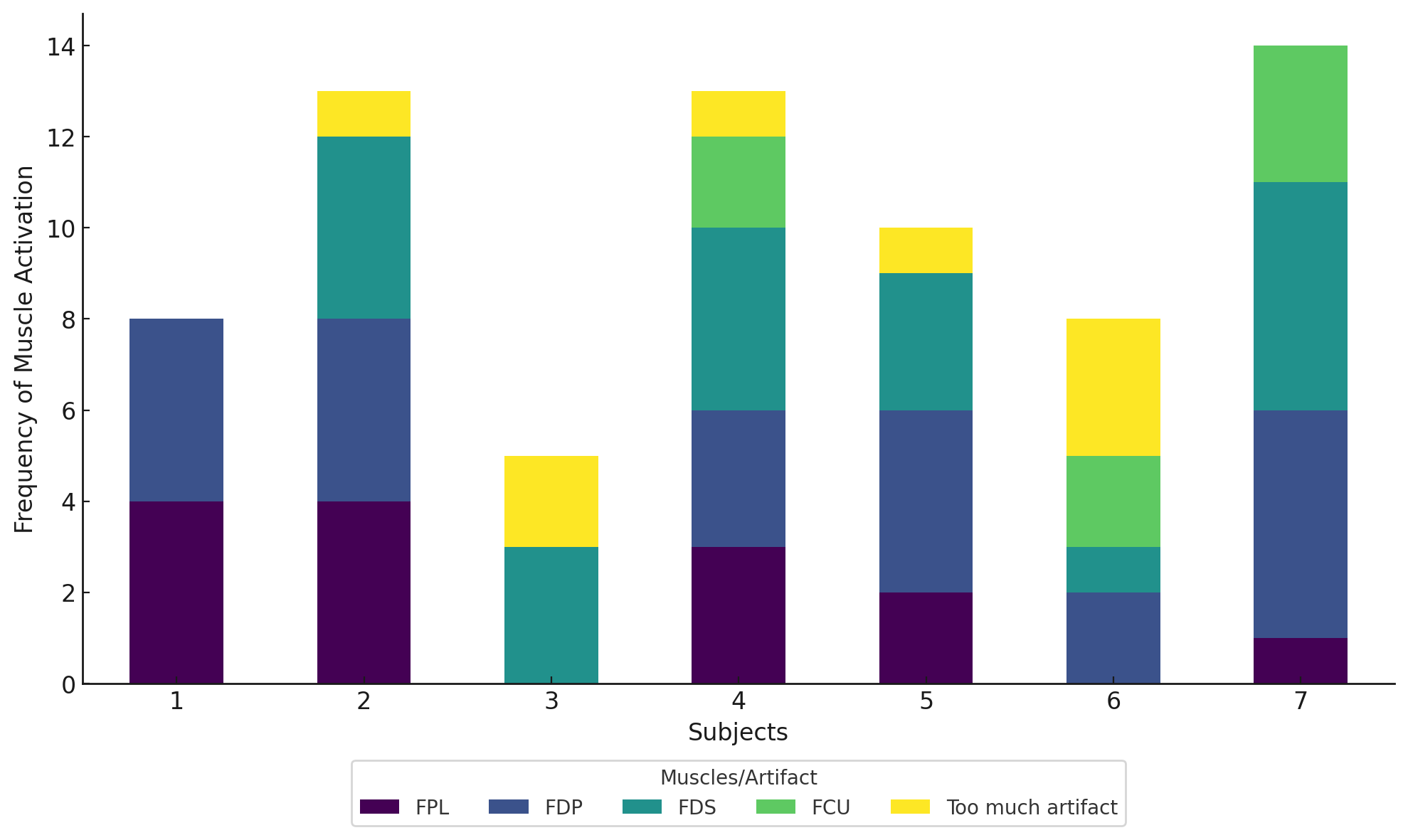}
    \caption{Subject-wise Grad-CAM based muscle activations}
    \label{fig:subject-gradcam}
\end{figure}

The Grad-CAM results were also analyzed to identify muscle activations across subjects. We hypothesize that FDP will exhibit consistent activation across subjects, with FPL and FDS also showing frequent activations, while subjects with higher artifacts will have less interpretable Grad-CAM outputs. The summarized results are presented in Table~\ref{tab:gradcam_results_subjects} and Figure \ref{fig:subject-gradcam}. 

\begin{table}[h!]
\centering
\caption{Frequency of muscle activations and artifacts across subjects.}
\label{tab:gradcam_results_subjects}
\begin{tabular}{|c|c|c|c|c|c|}
\hline
\textbf{Subject} & \textbf{FPL} & \textbf{FDP} & \textbf{FDS} & \textbf{FCU} & \textbf{Too much artifact} \\
\hline
1 & 3 & 5 & 0 & 0 & 0 \\
\hline
2 & 5 & 5 & 5 & 0 & 0 \\
\hline
3 & 0 & 0 & 5 & 0 & 5 \\
\hline
4 & 2 & 4 & 4 & 3 & 1 \\
\hline
5 & 2 & 4 & 4 & 0 & 1 \\
\hline
6 & 0 & 2 & 2 & 2 & 3 \\
\hline
7 & 1 & 5 & 5 & 2 & 0 \\
\hline
\end{tabular}
\end{table}
The results show that FDP was the most frequently activated muscle across subjects, showing consistent involvement in manipulation tasks. FPL and FDS were also commonly activated, with FDS showing significant activation in Subjects 2, 4, and 7. FCU showed relatively fewer activations, mainly for Subjects 4, 6, and 7. Subjects 3 and 6 exhibited the highest frequency of artifacts, limiting interpretability for these subjects. Subjects 1, 2, 4, 5, and 7 had relatively clean Grad-CAM outputs with fewer artifacts. These results validate the hypothesis that FDP is consistently activated across subjects, with artifacts limiting interpretability for specific subjects (e.g., Subjects 3 and 6).

\section{Discussion}\label{sec:discussion}
This section summarizes our conclusions and elaborates on their implications for skill classification, force estimation and interpretability. That is followed by a section outlining the limitations of our results and interpretation. 
\subsection{Skill Classification}
The classification model demonstrated strong performance in distinguishing manipulation skills using forearm ultrasound data, with an averaged cross-validation test accuracy of 94.9\% $\pm$ 10.2\% across subjects. These results highlight the model's ability to generalize well across diverse manipulation tasks and subjects.

The fold-wise analysis (Figure \ref{fig_fold_accuracy_stats} and Table \ref{table_fold_accuracy_stats}) shows consistently high training accuracies across all folds, indicating effective learning during training. However, test accuracy varies significantly between folds, with fold 2 exhibiting the lowest mean test accuracy (89.02\%) and highest variability ($\sigma_{test}$ = 15.98\%), suggesting possible challenges in this subset of the data. Fold 4 achieves the highest test accuracy (98.91\%) with minimal variability ($\sigma_{test}$ = 2.74\%), indicating that certain folds may align better with the model’s learned features. This variability underscores the importance of robust cross-validation to evaluate generalization performance comprehensively.

The subject-wise results (Figure \ref{fig:subject-classification} and Table \ref{subject-classification}) reveal high training accuracies (mean $\geq$ 99.92\%) with negligible variability, confirming stable model convergence during training. Test performance, however, varies across subjects. Subjects 2, 3, and 6 exhibit both high mean test accuracy and low variability (e.g., Subject 2: 99.99\% $\pm$ 0.04\%), indicating reliable generalization. Conversely, Subject 1 achieves the lowest test accuracy (88.10\%) with higher variability ($\sigma_{test}$ = 9.96\%), while Subject 5 shows the highest test variability ($\sigma_{test}$ = 18.59\%), suggesting individual subject variations.

\subsection{Force Estimation}
The force estimation results highlight the model's ability to generalize across diverse tasks and subjects using forearm ultrasound data. The model achieves an average test RMSE of 0.51 $\pm$ 0.19 N across all folds, subjects, and skills.

The fold-wise RMSE analysis (Figure \ref{fig:fold-regression} and Table \ref{table_fold_rmse_percent}) demonstrates consistently low training RMSE values (0.37 - 0.39 N) across all folds, corresponding to 9.24\% - 9.83\% of the 4 N sensor range. Testing RMSE shows greater variability, with Fold 5 yielding the highest test RMSE (0.58 N, 14.44\% of 4 N) and Fold 2 achieving the lowest (0.47 N, 11.69\% of 4 N). This variability suggests sensitivity to fold-specific data distributions, highlighting the need for balanced and representative cross-validation datasets.

Subject-wise analysis (Figure \ref{fig:subject-regression} and Table \ref{table_subject_rmse_percent}) reveals low training RMSE across subjects (0.29 - 0.46 N, 7.18\% - 11.54\% of 4 N), reflecting stable training performance. However, test RMSE varies significantly, with Subject 5 achieving the lowest test RMSE (0.34 N, 8.45\% of 4 N) and Subject 6 the highest (0.64 N, 16.11\% of 4 N). Variability in testing RMSE ($\sigma_{test}$) is highest for Subjects 4 and 6 (5.04 \% and 4.85 \% of 4 N, respectively), suggesting that these subjects' data may contain inconsistencies or unique patterns that challenge the model's generalization.

The skill-wise results (Figure~\ref{fig:skill-regression} and Table~\ref{table_class_rmse_percent}) show that Skill 3 demonstrates the highest RMSE values, both in training (0.54 N, 13.50\% of 4 N) and testing (0.68 N, 17.00\% of 4 N), indicating higher variability in data for this task. In contrast, Skill 2 consistently achieves the lowest RMSE (0.38 N in training and 0.46 N in testing), reflecting more straightforward force estimation for this task. Across all skills, testing RMSE and its variability are higher than their training counterparts, with testing standard deviation ranging from 0.12 N (3.00\% of 4 N, Skill 2) to 0.22 N (5.50\% of 4 N, Skill 3). This suggests greater heterogeneity in the testing dataset compared to the training dataset.

\subsection{Interpretability}
The Grad-CAM analysis provided valuable insights into the regions of ultrasound images associated with manipulation skill classification and force estimation, highlighting the model's interpretability. Across skills, the consistent activation of flexor digitorum profundus (FDP) highlights its role in tasks involving finger flexion. Similarly, frequent activations of flexor pollicis longus (FPL) and flexor digitorum superficialis (FDS) across multiple skills suggest their broader involvement in manipulation tasks. In contrast, the flexor carpi ulnaris (FCU) exhibited less frequent activations, likely reflecting its role in tasks requiring wrist stability rather than direct manipulation. The variability observed across skills, particularly the higher artifacts in Skills 3 and 5, underscores the complexity and potential challenges in interpreting ultrasound data for certain tasks. Skills 1 and 2 demonstrated consistent activations and minimal artifacts, making them more suitable for model-based predictions.

Subject-wise Grad-CAM analysis revealed the highest activation frequency for FDP, consistently involved across subjects. While FPL and FDS were commonly activated, subjects such as 2, 4, and 7 demonstrated stronger activations of FDS, potentially reflecting individual anatomical or task execution differences. Subjects 3 and 6 exhibited significant artifacts, limiting the interpretability of their results. This suggests variability in ultrasound image quality or differences in probe placement, muscle movement, or skin properties that may impact the robustness of the model's predictions.

Additionally, the occurrences of artifacts were noted in multiple subjects and skills; however, a quantitative assessment of artifact prevalence was not conducted. Future work will focus on defining objective metrics for quantifying artifacts and their impact on model interpretability. This will enable a clearer understanding of how artifacts affect classification performance and guide improvements in data acquisition and processing.

\subsection{Online Estimation}
\begin{figure}[h!]
    \centering
    \includegraphics[width=0.72\textwidth]{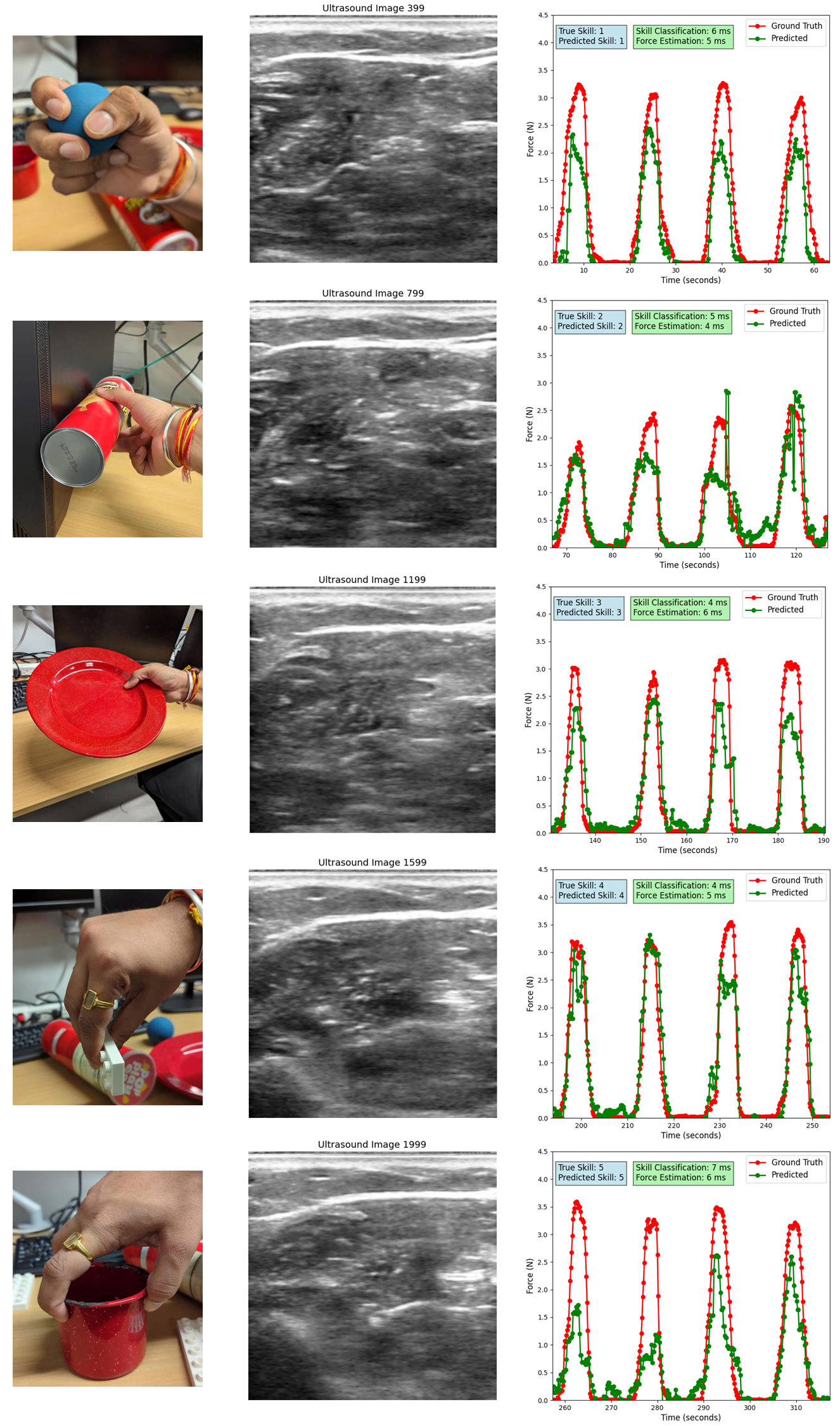}
    \caption{Ultrasound-Based Force Estimation and Skill Classification in Manipulation Tasks: Real-world tasks (left), corresponding ultrasound images (middle), and force estimation results (right) show ground truth (red) vs. predicted forces (green). Skill classification accuracy and inference times are overlaid for each task.}
    \label{fig:video}
\end{figure}

We provide an online estimation video in the supplementary section. For the video, test data from a randomly chosen subject was used to demonstrate the performance of the trained skill classification and force estimation models. The data were fed at the acquisition rate of 6.3 Hz. For our system, the average inference time for skill classification was 7.1 ms and for force estimation was 7.5 ms. This shows that our pipeline can be used at very high frame rates, which can be improved with faster and more efficient ultrasound image acquisition systems. Figure \ref{fig:video} shows the different manipulation skills and the corresponding ultrasound data and force estimation outputs.
% \textcolor{red}{Fill in here}

\subsection{Limitations and Future Work}
The limitations and future work for skill classification, force estimation and interpretability are described below.
\subsubsection{Skill Classification}
The variability in test accuracy across folds and subjects suggests sensitivity to data partitioning and individual differences, which could stem from factors such as anatomical variations, ultrasound signal quality, or task execution inconsistencies. While high accuracies indicate effective classification, the model's reliance on consistent data quality may limit its robustness in real-world applications. To improve generalizability, future work should focus on enhancing data diversity during training by including subjects with varied anatomies and skill execution styles. Data augmentation techniques and domain adaptation methods could further mitigate the impact of signal variability. Additionally, real-time validation in practical settings would help assess the model’s robustness and applicability to real-world robotic control systems.

Also, since each distinct object is involved in a distinct manipulation skill, there may be a strong correlation between finger postures and skill classification. This could limit generalizability in real-world scenarios where different skills may share similar finger configurations or where individuals adopt different postures for the same skill. Future work should investigate ways to decouple skill classification from finger posture dependency, such as incorporating temporal motion patterns or a broader task context.

\subsubsection{Force Estimation}
The variability in test RMSE across folds, subjects, and skills suggests the model’s sensitivity to the quality and consistency of the ultrasound data. Anatomical differences among subjects and inherent task complexity likely contribute to this variability. Additionally, the reliance on a fixed sensor range (4 N) may not generalize to applications requiring a broader force range. Improving the generalizability of force estimation models requires addressing data inconsistencies through techniques such as data augmentation and domain adaptation. Collecting a more diverse dataset with greater representation of anatomical and task variability could also enhance performance. Real-time testing in practical robotic manipulation tasks would validate the model's robustness and applicability for real-world scenarios.

An important aspect that remains to be analyzed is whether force estimation errors correlate with signal magnitude. If smaller forces exhibit proportionally higher errors, this could significantly impact tasks requiring precise low-force control. Future work should investigate error scaling across different force ranges to determine whether a normalization approach or adaptive modeling technique could mitigate potential biases toward larger force values.

\subsubsection{Interpretability}
While the Grad-CAM visualizations provide interpretability, their reliability depends on the quality of ultrasound data. Artifacts and variability across tasks and subjects highlight the model's sensitivity to input data quality. Differences in anatomical structure, probe placement, and task execution introduce inconsistencies that can hinder generalization. Additionally, the presence of artifacts, particularly for Skills 3 and 5, and Subjects 3 and 6, limits the scope of conclusions drawn from the results. The study assumes uniformity in image acquisition and preprocessing, which may not hold in practical, real-world scenarios. To improve the generalizability of the Grad-CAM-based insights, future work should focus on enhancing data quality by refining ultrasound acquisition protocols to minimize artifacts. Including a larger and more diverse subject pool can help address variability in anatomical features and task execution styles. Techniques such as artifact removal, domain adaptation, and improved probe stabilization could further enhance model robustness. Additionally, testing these insights in real-world robotic manipulation tasks would validate their practical applicability, ensuring that the Grad-CAM-guided interpretations can inform skill-specific training and decision-making in robotic systems. Expanding to more dynamic and complex manipulation skills could also explain the roles of additional muscle groups.

Regarding the occurrence of artifacts, a quantitative assessment of artifact prevalence was not conducted. Future work will focus on defining objective metrics for quantifying artifacts and their impact on model interpretability. This will enable a clearer understanding of how artifacts affect classification performance and guide improvements in data acquisition and processing techniques.

\section{Conclusions}
In this study, we demonstrated the use of a CNN-based framework for classifying manipulation skills and estimating forces using forearm ultrasound data. The proposed models achieved an average test accuracy of 94.9\% ± 10.2\% for skill classification and a test RMSE of 0.51 ± 0.19 N for force estimation across diverse subjects. Grad-CAM visualizations provided valuable insights into muscle activations, towards interpretability of the framework while highlighting the role of key muscle groups for different manipulation skills. Although variability was observed across folds, subjects, and tasks, the results demonstrate the capability of ultrasound imaging for simultaneous manipulation skill and force estimation for robotic teleoperation and training Learning from Demonstration models. This work demonstrates that ultrasound-based approaches can provide a reliable and interpretable means of understanding human muscle activity, towards improving human-machine interaction. Future efforts aimed at enhancing data diversity could expand the applicability of this framework. This can enable precise and adaptive robotic manipulation systems for industrial, healthcare, and assistive robotics scenarios.

% This study demonstrates the efficacy of a CNN-based framework for classifying manipulation skills and estimating force levels using forearm ultrasound data, achieving high accuracy across diverse tasks and subjects. The classification model achieved an average test accuracy of 94.9\% ± 10.2\%, and the force estimation model achieved a test RMSE of 0.51 ± 0.19 N, highlighting the potential of ultrasound-based approaches for robot teleoperation and training learning from demonstration models. Grad-CAM visualizations provided valuable insights into key muscle activations towards the interpretability of the proposed models. Variability in results across folds, subjects, and tasks underscores the need for more robust training protocols and data acquisition techniques to ensure consistency and generalizability. By addressing these limitations and expanding validation to real-world robotic manipulation scenarios, this approach could significantly enhance human-machine interaction systems in industrial, healthcare, and assistive robotics applications.

\section*{Acknowledgments}
\subsection*{Author Contributions} 
Conceptualization, K. B.; Funding acquisition, H. K. Z., R. D. H., and B. C.; Methodology, K. B., S. K., M. D.; Software, K. B., S. K.; Supervision, H. K. Z., R. D. H., and B. C.; Analysis, K. B., S. N., D. B. T.; Writing, K. B., S. N.

All authors have read and agreed to this version of the manuscript.

% Here are the author contributions and roles:

% K. Bimbraw conceived the idea and designed and conducted the experiments. 

% S. Nekkanti assisted with the data acquisition and the interpretability experiments. 

% D. B. Tiller assisted with the interpretability analysis.

% M. Deshmukh assisted with the defining the manipulation skills.

% B. Calli assisted with his manipulation expertise. 

% R. D. Howe assisted with his manipulation and robotics control expertise.

% H. K. Zhang assisted with his ultrasound expertise. 

\subsection*{Funding}
This work was supported by Amazon Robotics Greater Boston Technology Initiative (GBTI) grant and Worcester Polytechnic Institute (WPI) internal fund. 

\subsection*{Conflicts of Interest}
The author(s) declare(s) that there is no conflict of interest regarding the publication of this article.

\subsection*{Data Availability}
The data used to support the findings of this study are available from the first author upon request.

\section*{Supplementary Materials}
Movies S1 (Online Estimation) and S2 (Grad-CAM). 

% Please group supplementary materials in the following order: materials and methods, figures, tables, and other files (such as movies, data, interactive images, or database files). 

% \medskip Example:
% Fig. S1. Title of the first supplementary figure.

% Fig. S2. Title of the second supplementary figure.

% Table S1. Title of the first supplementary table.

% Data file S1. Title of the first supplementary data file.

% Movie S1. Title of the first supplementary movie.

% \medskip
% Be sure to submit all supplementary materials with the manuscript and remember to reference the supplementary materials at appropriate points within the manuscript. We recommend citing specific items, rather than referring to the supplementary materials in general, for example: ``See Figures S1-S10 in the Supplementary Material for comprehensive image analysis.''

% A link to access the supplementary materials will be provided in the published article.

% Supplementary Materials may include additional author notes—for example, a list of group authors.

\end{document}